%% file: main.tex
\documentclass[10pt,twocolumn,letterpaper]{article}

\usepackage[pagenumbers]{cvpr} 

\input{preamble}

\definecolor{cvprblue}{rgb}{0.21,0.49,0.74}
\usepackage[pagebackref,breaklinks,colorlinks,citecolor=cvprblue]{hyperref}

\usepackage{multirow}
\usepackage{bbding}

\usepackage{diagbox}
\def\paperID{8213} 
\def\confName{CVPR}
\def\confYear{2024}

\title{Compensation Sampling for Improved Convergence in Diffusion Models}

\author{%
    Hui Lu\textsuperscript{1}, Albert ali Salah\textsuperscript{1, 2}, Ronald Poppe\textsuperscript{1}\\
    $^{1}$ Utrecht University, $^{2}$Bogazici University\\
    \texttt{\{h.lu1, a.a.salah, r.w.poppe\}@uu.nl}
}

\begin{document}

\maketitle
\begin{abstract}
Diffusion models achieve remarkable quality in image generation, but at a cost. Iterative denoising requires many time steps to produce high fidelity images. We argue that the denoising process is crucially limited by an accumulation of the reconstruction error due to an initial inaccurate reconstruction of the target data. This leads to lower quality outputs, and slower convergence. To address this issue, we propose \textit{compensation sampling} to guide the generation towards the target domain. We introduce a compensation term, implemented as a U-Net, which adds negligible computation overhead during training and, optionally, inference. Our approach is flexible and we demonstrate its application in unconditional generation, face inpainting, and face de-occlusion using benchmark datasets CIFAR-10, CelebA, CelebA-HQ, FFHQ-256, and FSG. Our approach consistently yields state-of-the-art results in terms of image quality, while accelerating the denoising process to converge during training by up to an order of magnitude.
\end{abstract}

\section{Introduction}
Diffusion models have achieved great success in image generation~\cite{ho2020denoising,lugmayr2022repaint,pandey2022diffusevae,preechakul2022diffusion}. Compared to other deep generative models such as Generative Adversarial Networks (GANs)~\cite{dhariwal2021diffusion,NIPS2014_5ca3e9b1}, diffusion models offer stable training, easy model scaling, and good distribution coverage~\cite{moghadam2023morphology}. But despite their success, diffusion models suffer from low training and inference efficiency because they typically require many time steps to converge and to generate high-quality outputs from the denoising process. Compared to GANs, which only require a single forward pass through the generator network, inference in diffusion models is two to three orders of magnitude slower~\cite{song2021denoising}. Simply reducing the number of time steps disrupts the Gaussian assumption of the denoising process, and has been shown to reduce the synthesis quality~\cite{sohl2015deep,xiao2021tackling}.

\begin{figure}
\centering
\includegraphics[width=\columnwidth]{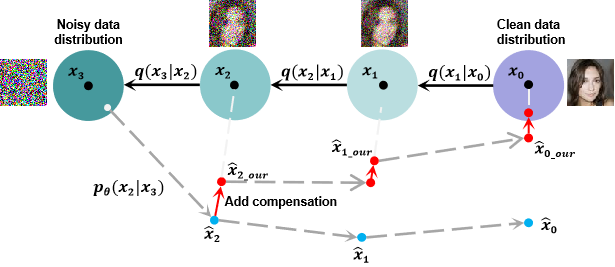}
\caption{\textbf{Compensation sampling (red) compared to traditional sampling (dashed).} Both processes start from a noisy data distribution. Compensation sampling guides the reconstruction towards the clean data distribution for faster convergence. \vspace{-2mm}}
\label{fig:1}
\end{figure}

Due to the iterative sampling, inaccurate reconstruction early in the training process causes the accumulation of reconstruction errors in subsequent time steps~\cite{liu2022flow}. This hinders convergence speed and the final quality of the model. To address the issue of error accumulation, several works have encoded conditionals in the denoising process to improve the initial sample quality, and to speed up the process. These approaches include encoding image features or images generated by Variational Autoencoders (VAE)~\cite{pandey2022diffusevae,preechakul2022diffusion}, and assigning prioritized weights on specific noise levels in the denoising process~\cite{choi2022perception,daras2023soft}.

Conditioned denoising processes take fewer time steps to converge but the additional conditions often require significant computation cost. Conditioning is also arguably more dataset-specific, which could limit generalization. But more importantly, simply adding conditions does not address inherent issues in the denoising process. In other words, it does not avoid the error accumulation that causes slower convergence towards a sub-optimal model. The challenge thus remains to improve the efficiency during training and inference in a principled way without compromising the output quality. This is especially true for unconditional generation, in the absence of a powerful conditioning signal. 

In this work, we address the efficiency issue of diffusion models by introducing \textit{compensation sampling} (CS). It allows us to use fewer time steps during training and inference without breaking the Gaussian assumption of the denoising process. Compensation sampling can be applied in both unconditional and conditional generation tasks. An illustration of compensation sampling appears in Figure~\ref{fig:1}. At the core of our approach is the use of a learned compensation term to direct the reconstruction towards the clean data distribution, and consequently avoid error accumulation. We show that this process results in quicker training convergence, and higher-quality images. Our main contributions are:
\begin{enumerate}
    \item \textbf{Novel sampling algorithm}. We propose the compensation sampling algorithm, with rigid mathematical derivation, that can reduce the number of time steps in diffusion models during training by an order of magnitude.

    \item \textbf{State-of-the-art results}. We apply compensation sampling and achieve results that are on par with, and typically outperform, current state-of-the-art diffusion models on unconditional generation, face inpainting, and face de-occlusion on benchmark datasets CIFAR-10, CelebA, CelebA-HQ, FFHQ, and FSG.
\end{enumerate}

We first discuss related works for accelerating the convergence, and then detail our method in Section~\ref{sec: our method}. We continue with a presentation of our experiments and ablation study in Section~\ref{sec:experiments}, and finally conclude in Section~\ref{sec:conclusions}.

\section{Related Work}
\label{sec:related work}

Denoising diffusion probabilistic models (DDPM)~\cite{ho2020denoising} are generative models that achieve state-of-the-art generation performance. They have several advantages over other generative models such as Generative Adversarial Networks (GAN) ~\cite{dhariwal2021diffusion,NIPS2014_5ca3e9b1,karras2020analyzing} and Variational Autoencoders (VAE)~\cite{vahdat2020nvae}. Diffusion models transform a complex clean data distribution $p_{data}(x)$ into noise distribution $\mathcal N(0, I)$ and learn the reverse process to restore data from noise. Based on DDPM, Denoising Diffusion Implicit Models (DDIM)~\cite{song2019generative,song2020score} have been proposed. They have an efficient sampling process that has the same training objectives as DDPM: an iterative approach to solve the stochastic differential equation. 
Although DDIM shows impressive image generation results with few time steps, during the training process, it takes many time steps (e.g., $T = 1,000$) for the sampling process to fully converge. To make the process more efficient, researchers have reconsidered the sampling process during training. Because the sampling process starts from the image with low quality in early training, the sampling error will accumulate when sampling at subsequent time steps~\cite{liu2022flow}. This causes the training to converge slower, and reduces the final image quality.

Several works have addressed this issue. In Soft Diffusion~\cite{daras2023soft}, a Momentum Sampler is proposed to generate the samples under different types of corruptions, i.e., types of noise. \citet{Meng_2023_CVPR} employ a second model to match the output of the combined conditional and unconditional diffusion models, and then distills that model to accelerate the sampling process. \citet{fei2023generative} use a protocol of conditional guidance, which enables the diffusion models to generate images of arbitrary resolutions. \citet{xu2023stable} combine the reference batch to reduce the covariance, such that they can accelerate the intermediate regime generation. \citet{Li_2023_ICCV} use an additional PTQ tool to compresses the noise estimation network to accelerate the generation process.

One specific approach to improve training convergence in diffusion models is by using additional conditionals during the sampling process. \citet{preechakul2022diffusion} propose a semantic encoder to encode image features into the sampling process during training to accelerate the inference while generating images with high quality. Similarly, \citet{pandey2022diffusevae} inject VAE-generated images into the sampling process as additional conditions during training. 
\citet{choi2022perception} prioritize the later noise levels. An emphasis on the training of the reverse stage of these later noise levels encourages the model to reduce the accumulated error. Consequently, the convergence is sped up while better image quality is typically obtained.



In this paper, we depart from the idea of encoding additional conditionals into the sampling process during training. We propose a compensation algorithm for diffusion models that not only boosts the convergence during training without breaking any of the assumptions, but also improves the image quality. Our approach can be applied in various generation tasks, in both unconditional and conditional generation. Closest to our work is the recently introduced Cold Diffusion~\cite{bansal2022cold}. This work uses a Taylor expansion of the degradation function to represent the diffusion process, but it has to deal with diffusions that lack significant gradient information. While the method is beneficial for linear degradations, it fails to improve over the DDIM baseline because it violates the linear degradation assumption due to the addition of Gaussian noise. In contrast, our proposed approach has rigid mathematical underpinnings that adhere to the Gaussian assumption of the denoising process.

\begin{figure*}[htb]
\centering
\includegraphics[width=\textwidth]{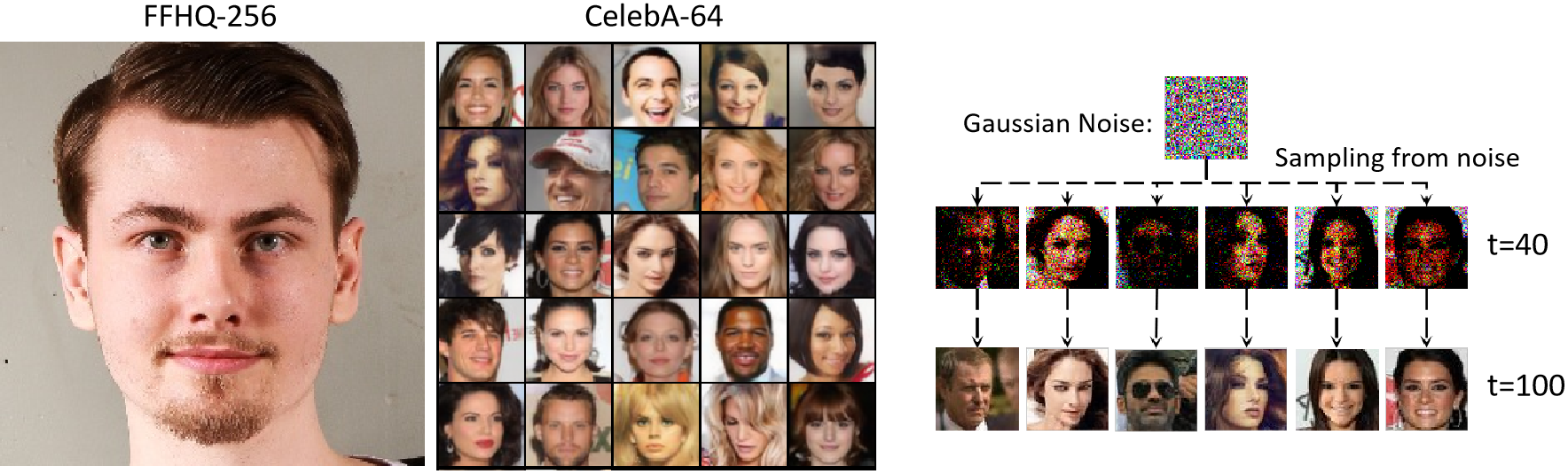} \vspace{-8mm}
\caption{\textbf{Visualization of our compensation diffusion results.} (left) The unconditional generation results of FFHQ-256 ($256 \times 256$ resolution) and CelebA-64 ($64 \times 64$ resolution) datasets (NFE = 100). (right) Illustration of the diversity of our compensation sampling on random Gaussian noise, visualized at time steps 40 and 100.
}
\label{fig:2}
\end{figure*}

\section{Compensation diffusion}
\label{sec: our method}
We discuss common sampling practice in diffusion models before introducing our compensation sampling. We then discuss the architecture and training of the model with compensation sampling that is used in our experiments.

\subsection{Common sampling in diffusion models}
\label{subsec:1}
Diffusion models first degrade a complex clean data distribution $p_{data}(x)$ into noise distribution $\mathcal N(0, I)$ and learn the reverse reconstruction process to restore data from noise~\cite{ho2020denoising}. The diffusion process gradually corrupts clean data $x_0$ with predefined noise scales $0 < \beta_1 < \beta_2, ..., \beta_T < 1$, indexed by time step $t$ ($1 \leq t \leq T$), and $T$ the number of time steps. Corrupted data $x_1, ..., x_T$ are sampled from $x_0 \sim p_{data}(x)$ with a diffusion process, defined as the Gaussian transition:

\begin{equation}
q(x_t|x_{t-1}) = \mathcal N(x_t;\sqrt{1-\beta_t}x_{t-1}, \beta_t\rm{I})
\label{equa:1}
\end{equation}

When a clean data point $x_0$ is provided, the noisy $x_t$ can be obtained from the following equation:
\begin{equation}
q(x_t|x_0) = \mathcal N(x_t;\sqrt{\overline\alpha_t}x_0, (1-\overline\alpha_t)\rm{I})
\label{equa:2}
\end{equation}
where $\alpha_t = 1 - \beta_t$ and $\overline\alpha_t = \prod_{s=1}^{t}\alpha_s$. When $t$ is close to $T$, $x_t$ can be approximated as a Gaussian distribution.

Diffusion models learn the reverse process to generate samples from the data distribution. The optimization objective of the reverse transition can be derived from a variational bound~\cite{kingma2013auto}. \citet{ho2020denoising} employ a variational solution and assume its reverse transition kernel also subjects to a Gaussian distribution. This way, the generation process parameterizes the mean of the Gaussian transition distribution and fixes its variance as~\cite{li2022entropy}:
\begin{eqnarray}
p_\theta(x_{t-1}|x_t)=\mathcal N(x_{t-1};\mu_{\theta}(x_t),\sigma_t^2\rm{I})
\label{equa:3}\\
\mu_{\theta}(x_t)= \frac{1}{\sqrt{\alpha_t}}(x_t-\frac{1-\alpha_t}{\sqrt{1-\overline\alpha_t}}\epsilon_{\theta}(x_t))
\label{equa:4}
\end{eqnarray}

\noindent with noise estimator $\epsilon_{\theta}(x_t)$ typically modeled by a neural network with variance $\sigma_t^2$ as a training hyper-parameter.

The authors of DDIM~\cite{song2021denoising} propose an efficient sampling process that has the same training objectives as DDPM: an iterative approach to solve the stochastic differential equation, which can be expressed as:

\begin{equation}
\begin{split}
    x_{t-1}&=\sqrt{\alpha_{t-1}}f_{\theta}(x_t,t) + \sqrt{1-\alpha_{t-1}-\sigma_t^2}\epsilon_{\theta}(x_t)\\
    &\quad+\sigma_t^2\rm{z}
    \label{equa:5}
\end{split}
\end{equation}
where $f_{\theta}(x_t,t)$ is the prediction of clean data point $x_0$ when the noisy $x_t$ is observed and noise prediction $\epsilon_{\theta}(x_t)$ is given. $f_{\theta}(x_t,t)$ can be expressed as:
\begin{equation}
    f_{\theta}(x_t,t) = \frac{x_t-\sqrt{1-\alpha_t}\epsilon_{\theta}(x_t)}{\sqrt{\alpha_t}}
\label{equa:6}
\end{equation}

When variance $\sigma_t$ is set to 0, the sampling process becomes deterministic. The non-Markovian diffusion process \cite{song2021denoising} allows the generation quality to remain unchanged with fewer denoising steps.

In Equations~\ref{equa:3}-\ref{equa:6}, a neural network is typically used to predict the original data point $x_0$ to obtain a sample distribution $p_\theta(x_{t-1}|x_t)$. However, since the prediction of $x_0$ is inaccurate at the start of training, convergence will be slow. In the conditional sampling algorithm, the solution is to iteratively predict the original data $x_0$ based on $x_{t-1}$~\cite{ramesh2022hierarchical,xiao2021tackling}:
\begin{equation}
    p_\theta(x_{t-1}|x_t)=q(x_{t-1}|x_t,x_0 \approx \hat{x_0}=f_{\theta}(x_t,t))
    \label{equa:7}
\end{equation}
First, $x_{t-1}$ is sampled using the posterior distribution in Equation~\ref{equa:7}, and subsequently used as the input to predict the original data $x_0$. The above sampling process iteratively runs over $t$ until the final result $x_0$ is generated.

\subsection{Limitation of common sampling}
We first consider deterministic sampling where the noise pattern is selected at the start of the generation process. The diffusion process can be mathematically presented as:
\begin{equation}
    D(x,t) = \sqrt{1-\beta_t}x + \sqrt{\beta_t}z
    \label{equa:8}
\end{equation}
with $D(x,t)$ the deterministic interpolation between data $x$ and fixed noise pattern $z \in \mathcal N(0, I)$, at time step $t$.

To obtain sample distribution $p_\theta(x_{t-1}|x_t)$, a neural network is typically used to predict original data $x_0$, or noise $\epsilon$. During the iterative denoising process, when the optimal reconstruction model $R$ is learned, i.e., when $R(D(x_0, t), t) = x_0$ for all time steps $t$, the conditional sampling produces exact iterations $x_t = D(x_0, t)$. 
However, during the training process, obtaining the perfect reconstruction model $R$ is far from trivial. Especially in early training stages, iterations $x_t$ tend to move away from $D(x_0, t)$. The errors caused by the inaccurate reconstruction will accumulate over the time steps since the sampling process is iterative. This will lead to an extended convergence time, and reduced generated image quality.


\subsection{Compensation sampling}
To prevent reconstruction error accumulation, we propose the compensation sampling (CS) algorithm. Based on Equation~\ref{equa:8}, we define a general deterministic diffusion process:
\begin{equation}
    D(x,t) = g(t)x + f(t)z
    \label{equa:9}
\end{equation}
where $g(t)$ and $f(t)$ are functions that define the noise scales, and $z \in \mathcal N(0, I)$. Considering the forward diffusion process in deterministic sampling, we can define $x_t$ as:
\begin{equation}
    x_t = g(t)x_0 + f(t)z
    \label{equa:new1}
\end{equation}

The compensation sampling algorithm relies on an initial sample reconstruction:
\begin{equation}
   \hat{x_0} = R(x_t,t)
   \label{equa:10}
\end{equation}

With $\hat{x_0}$, we define the compensation weight $w(t)$ as:

\begin{equation}
    w(t) = g(t) - g(t-1)
    \label{equa:11}
\end{equation}

We then obtain the perfect inverse of $x_{t-1}$ from $x_t$:

\begin{equation}
\begin{split}
x_{t-1}&= x_t - D(\hat{x_0},t) + D(\hat{x_0},t-1)\\
&\quad+ w(t)(\hat{x_0}-x_0) \\
&=g(t)x_0 -g(t)\hat{x_0}+ g(t-1)\hat{x_0}+f(t-1)z\\
&\quad + w(t)(\hat{x_0}-x_0)\\
&=\hat{x_0}(g(t-1)-g(t))+ w(t)\hat{x_0} + g(t)x_0 \\
&\quad- w(t)x_0 +f(t-1)z\\
&=g(t)x_0 - w(t)x_0+f(t-1)z\\
&=g(t-1)x_0 + f(t-1)z\\
&=D(x_0,t-1)
\end{split}
\label{equa:12}
\end{equation}

From Equations~\ref{equa:9}-\ref{equa:12}, it follows that $x_t = D(x_0, t)$ for all $t < T$, regardless of $R$. This means that the accumulated error will be highly alleviated since the iterates $x_t$ will be the same as when $R$ is a perfect inverse for the degradation $D$. We notice that Cold Diffusion~\cite{bansal2022cold} also adds the term $D(\hat{x_0},t) + D(\hat{x_0},t-1)$ during sampling. However, only using this term shows worse performance than DDIM~\cite{song2021denoising}. Instead, our algorithm mathematically realizes the perfect inverse to reduce the accumulated error. We define $w(t)(\hat{x_0}-x_0)$ as the \textit{compensation term}. Our ablation studies provide insight into the practical operation of the term during training.

We use the final image $x_0$ in Equation~\ref{equa:12} but, during generation, we do not have access to ground truth $x_0$. Consequently, we cannot directly calculate the compensation term. Also, using the perfect inverse will reduce the ability to generate diverse outputs. To eliminate these issues, we utilize a U-Net model~\cite{ronneberger2015u} with negligible computation cost during training as a compensation module to learn the compensation term.  
Only a single training epoch is used, which means that the compensation module is barely trained, but the direction of the denoising process will still be guided towards the original data distribution. As an additional benefit, this approach will introduce some noise into the denoising process and help retain the diversity of diffusion models, see Figure~\ref{fig:2} (right) for some examples. In the supplementary material, we further show the diversity of the output of our approach and experiment with different numbers of training epochs for the compensation module.

With Equations~\ref{equa:9}-\ref{equa:12}, compensation sampling realizes improved convergence of diffusion models during training. Importantly, our approach is not limited to fixed noise scales $\beta_1, \beta_2, ..., \beta_T$ or fixed Gaussian noise patterns, but can be applied to any diffusion pattern. 

\subsection{Compensation diffusion model and training}
To make a fair comparison with recent diffusion models with common sampling, we opt to use the popular Ablated Diffusion Model (ADM)~\cite{dhariwal2021diffusion} as a backbone. ADM is based on the U-Net architecture~\cite{ronneberger2015u} with residual blocks and self-attention layers in the low-resolution feature maps. In our main experiments, we use our compensation sampling in DDIM~\cite{song2021denoising}, and term the resulting model DDIM+CS. For the compensation module, we use an additional lightweight U-Net model. See supplementary material for the architecture details.
For training, we use the inner iteration training scheme~\cite{such2020generative}. During a single training, the compensation module can be trained multiple times (default as one, see supplementary material for other options) using $L_1$ loss. With ADM, we obtain restoration model $R$. Since $R$ can be interpreted as an equally weighted sequence of denoising modules $\epsilon_{\theta}(x_t,t)$, we train by optimizing loss function $L_{DM}$ with respect to $\theta$:
\begin{equation}
    L_{DM}= \sum_{t=1}^T \mathbb{E}_{x_0,\epsilon_t}\left[ \left \|  \epsilon_{\theta}(x_t,t) -\epsilon_t  \right \|_2^2     \right]
\label{equa:13}
\end{equation}
with $x_0$ the original image, $\epsilon_t \in \mathbb{R}^{3 \times H \times W} \sim \mathcal N(0, I)$, and $T$ is the number of time steps.

\section{Experiments} \label{sec:experiments}
We experiment on benchmark datasets CIFAR-10~\cite{krizhevsky2009cifar10}, CelebA~\cite{liu2015deep}, CelebA-HQ~\cite{karras2018progressive}, FFHQ~\cite{karras2019style}, and FSG~\cite{cheung2021facial}. We address unconditional generation, conditional face inpainting, and face de-occlusion. We then analyze the training speed-up and present our ablation study. Additional results appear in the supplementary material.

\begin{table*} [t]
\centering
\resizebox{1.0\linewidth}{!}{
\begin{tabular}{|l|c|c|c|c|c|c|c|c|c|c|}
\hline
& &\multicolumn{5}{c|}{\textbf{CelebA-64}} & \multicolumn{3}{c|}{\textbf{FFHQ-256}}  \\
\hline
\textbf{Methods} &\diagbox{\textbf{Backbone}}{\textbf{NFE}} & 10 & 20 & 50 & 100  & 1,000 &  100 & 200 &500\\
\hline
Cold Diffusion~\cite{bansal2022cold} & ADM & 32.81 & 20.45 & 12.93 & 7.13 & 7.84 & 31.32 & 29.53 & 28.11 \\
DDIM~\cite{song2021denoising}& ADM & 17.33 & 13.73 &9.17 &6.53 &4.88 &  17.89 & 11.26 & 8.41 \\
\hline
P2 Diffusion~\cite{choi2022perception} & ADM & 20.37 & 16.11 & 11.96 & 9.04 & 7.22 & 16.78 & 10.38 & 6.97  \\
D2C~\cite{sinha2021d2c} & NVAE+U-Net & 17.32 & 11.46 & 6.80 & 5.70 & 5.15 &  13.04 & 9.85 & 7.94 \\
Diffusion Autoencoder~\cite{preechakul2022diffusion} & ADM & 12.92 & 10.18 & 7.05 & 5.30 & 4.97 & 15.33 & 8.80 & 5.81 \\
Analytic-DDIM~\cite{bao2022analyticdpm} & ADM & 15.62 & 10.45 & 6.13 & 4.29 & 3.13 & 14.84 & 9.02 & 5.98  \\
DPM-Solver~\cite{lu2022dpm} & ADM & 5.83 & 3.13 & 3.10 & 3.11 &3.08 & 10.82 & 8.47 & 8.40 \\
SN-DDIM~\cite{bao2022estimating} & ADM & 10.20 & 6.77 & 3.83 & 3.04 & 2.90 & 14.82 & 8.79 & 5.44  \\
F-PNDM~\cite{liu2022pseudo} & ADM & 7.71 & 5.51 & 3.34 & 2.81 & 2.86 & 17.51 & 8.23 & 4.79 \\

DDIM+CS (\textbf{ours}) & ADM & 7.80 & 5.11 & 2.23 & 2.11 & 1.98 & 11.89& 7.31 & 4.02 \\
DPM-Solver+CS (\textbf{ours}) ~\cite{lu2022dpm} & ADM & \underline{5.22} & \underline{2.35} & 2.22 & 2.12 & 2.04 & \underline{9.73} & \underline{4.66} &4.01\\

\hline
PDM~\cite{wang2023patch} (patch-wise training) & ADM &35.88 &8.36 & \underline{1.77} & \underline{1.86} & \underline{1.82} & 23.55 & 6.47 & \underline{3.13}\\
PDM+CS (\textbf{ours}) (patch-wise training) & ADM & \textbf{4.93} & \textbf{1.97} &  \textbf{1.42} & \textbf{1.44} & \textbf{1.38} & \textbf{6.11} & \textbf{3.52} & \textbf{2.57}\\

\hline
U-Net GAN ~\cite{schonfeld2020u} & U-Net &  \multicolumn{5}{c|}{19.31} & \multicolumn{3}{c|}{10.90} \\
VQGAN~\cite{esser2021taming} & CNN+Transformer & \multicolumn{5}{c|}{12.70} & \multicolumn{3}{c|}{9.60} \\
GANFormer2~\cite{arad2021compositional} & CNN+Transformer & \multicolumn{5}{c|}{6.87} & \multicolumn{3}{c|}{7.77} \\
Diffusion StyleGAN2 ~\cite{wang2023diffusiongan} & StyleGAN & \multicolumn{5}{c|}{\underline{1.69}} & \multicolumn{3}{c|}{\underline{3.73}} \\
Diffusion StyleGAN2+CS (\textbf{ours}) ~\cite{wang2023diffusiongan} & StyleGAN & \multicolumn{5}{c|}{\textbf{1.21}} & \multicolumn{3}{c|}{\textbf{2.95}} \\
\hline
\end{tabular}
}
\caption{\textbf{Unconditional image generation results}. Comparison on FID-50k with the state-of-the-art on CelebA-64 and FFHQ-256. Top part of the table contains baselines DDIM and Cold Diffusion, the middle part are state-of-the-art diffusion models, and the bottom part summarizes the performance of GAN models. Best results for diffusion and GAN models in \textbf{bold}, second best are \underline{underlined}. \vspace{-2mm}}
\label{tab:results_unconditional}
\end{table*}

\subsection{Unconditional generation}
The unconditional generation task aims to generate images without any conditioning or constraints.

\textbf{Experiment setting}. We evaluate on CIFAR-10 (60k images, $32 \times 32$ resolution), CelebA-64 (200k images, $64 \times 64$), and FFHQ-256 (70k images, $256 \times 256$). CIFAR-10 contains images of 10 different classes such as airplane and cat. CelebA-64 and FFHQ-256 are frontal face datasets. For fair comparison, we use the training hyper-parameters from DDIM~\cite{song2021denoising}. We use Gaussian noise as our corruption mechanism and adopt the fixed linear variance schedule $\beta_1, ..., \beta_T$ as in previous works~\cite{ho2020denoising,nash2021generating,song2021denoising,karras2022elucidating} for the diffusion process in Equation~\ref{equa:1}. Importantly, given the quicker convergence of compensation sampling, we reduce the training time steps $T$ to 100. Both compensation and diffusion model use the Adam optimizer.

\textbf{Evaluation metrics}. Ways to quantitatively assess how accurately the generated distributions mimic the training data distribution remain open research topics. We employ the widely used FID~\cite{heusel2017gans}. The FID score measures the KL divergence between two Gaussian distributions in the Inception-V3 feature space, computed by comparing real reference samples to generated samples. We randomly generate 50k images to compute the FID score (FID-50k) with the same implementation as in EDM~\cite{karras2022elucidating}.

\textbf{Quantitative evaluation}. In our comparisons, we report the performance from the published papers. We calculate missing scores when source code is provided by the authors.

\textbf{CelebA \& FFHQ}. We focus on face synthesis using diffusion models and GANs in Table~\ref{tab:results_unconditional}. We compare the reported FID-50k performance from the published papers with the number of function evaluations (NFE) during generation. The top part of the table contains diffusion-based methods including our baselines Cold Diffusion~\cite{bansal2022cold} and DDIM~\cite{song2020score} with common sampling, whereas the bottom part summarizes the performance of GANs.

We first discuss regular training, before moving to patch-based training. With regular training at NFE=1,000, our DDIM+CS achieves the best performance of all tested diffusion models. On CelebA, we outperform baselines DDIM with common sampling and Cold Diffusion by 59\% (4.88$\rightarrow$1.98) and 75\% (7.84$\rightarrow$1.98). For FFHQ-256 at NFE=500, gains of 52\% and 86\% are achieved. More importantly, all tested models at NFE=1,000 are also outperformed by DDIM+CS with NFE=100 or even NFE=50. This indicates that we obtain higher quality images with much lower training time. We analyze this in Section~\ref{sec:exp_computation_cost}.

We also obtain better performance at NFE=50, 100 and 1,000. At NFE=20, DPM-Solver has a lower FID-50k score. But when we apply our compensation sampling in DPM-Solver (DPM-Solver+CS), we obtain improved performance on both CelebA-64 and FFHQ-256.

Patch-wise training as used in PDM~\cite{wang2023patch} allows for even better results. Our resulting PDM+CS obtains state-of-the-art performance on both CelebA-64 (FID-50k=1.38 at NFE=1,000) and FFHQ-256 (FID-50k=2.57 at NFE=500). 

For GANs, the superior performance of Diffusion StyleGAN2 can be attributed to the powerful backbone. When applying our compensation diffusion method to this backbone (Diffusion StyleGAN2+CS), we obtain the best performance with an improvement of 28\% (1.69$\rightarrow$1.21) on CelebA-64 and 21\% (3.73$\rightarrow$2.95) on FFHQ-256.

\begin{table} [htb]
\centering
\resizebox{1.0\linewidth}{!}{
\begin{tabular}{|l|c|}
\hline
\textbf{Method} & \textbf{FID-50k} $\downarrow$ \\
\hline
DDPM (T=1,000, NFE=1,000)~\cite{ho2020denoising} & 3.17 \\
DDIM (T=1,000, NFE=1,000)~\cite{song2021denoising} & 3.95 \\
DiffuseVAE-72M (T=1,000, NFE=1,000)~\cite{pandey2022diffusevae} & 2.62 \\
DDPM++ (T=1,000, NFE=1,000)~\cite{kim2021soft} & 2.56 \\
LSGM (T=1,000, NFE=138)~\cite{vahdat2021score} & 2.10 \\
EDM (T=1,000, NFE=35)~\cite{karras2022elucidating} & 1.97 \\
PFGM++ (T=1,000, NFE=35)~\cite{xu2023pfgm++} &1.91 \\
\hline
DDIM+CS (\textbf{ours}) (T=100, NFE=35) & 2.01 \\
DDIM+CS (\textbf{ours}) (T=1,000, NFE=35) & \underline{1.57} \\
PFGM++ + CS (\textbf{ours}) (T=100, NFE=35) & 1.74 \\
PFGM++ + CS (\textbf{ours}) (T=1,000, NFE=35) & \textbf{1.50} \\
\hline
\end{tabular}
}
\caption{\textbf{Unconditional image generation results on CIFAR-10} using FID-50k.}
\label{tab:cifar}
\end{table}

\textbf{CIFAR-10}. To validate our approach on general image synthesis, we report on CIFAR-10 in Table~\ref{tab:cifar}. We evaluate with 35 NFE, in line with EDM~\cite{karras2022elucidating} and PFGM++~\cite{xu2023pfgm++}. Compared to approaches trained with 10 times more time steps, our DDIM+CS achieves competitive results with much less training time (see Table~\ref{tab:computation cost}). When trained with $T=1,000$, DDIM+CS (FID-50k=1.57) outperforms the current state-of-the-art. We observe that the previous best approach PFGM++~\cite{xu2023pfgm++} has a significantly more complex backbone, as it combines diffusion models and Poisson Flow Generative Models. When we apply compensation sampling in PFGM++ (PFGM++ + CS), a lower FID score is achieved (1.91$\rightarrow$1.74) with the same NFE, but 10 times fewer time steps. When we increase the time steps to $T=1,000$, we obtain the state-of-the-art result of 1.50.

\begin{table*} [htb]
\centering
\resizebox{1.0\linewidth}{!}{
\begin{tabular}{|l|c c|c c|c c|c c|c c|}
\hline
\multirow{2}{*}{\textbf{Methods}} & \multicolumn{2}{c|}{\textbf{Half}} &\multicolumn{2}{c|}{\textbf{Completion}} & \multicolumn{2}{c|}{\textbf{Expand}} & \multicolumn{2}{c|}{\textbf{Thick Line}}  & \multicolumn{2}{c|}{\textbf{Medium Line}} \\
& LPIPS $\downarrow$ & FID $\downarrow$  & LPIPS $\downarrow$ & FID $\downarrow$ & LPIPS $\downarrow$ & FID $\downarrow$ & LPIPS $\downarrow$ & FID $\downarrow$ & LPIPS $\downarrow$ & FID $\downarrow$  \\
\hline
CoModGAN~\cite{zhao2021large} & 0.445 & 37.72 & 0.406  & 43.77 & 0.671 & 93.48 &0.091 & 5.82 &0.105 & 5.86\\
LaMa~\cite{suvorov2022resolution} & 0.342 & 33.82 & 0.315 & 25.72 & 0.538 & 86.21 & 0.080  & 5.47  & 0.077 & 5.18 \\
CDE~\cite{batzolis2021conditional} & 0.344 & 29.33 & 0.302  & 19.07  & 0.508 & 71.99 & 0.079  & 4.77 & 0.070 &4.33 \\
RePaint~\cite{lugmayr2022repaint} & 0.435 & 41.28 & 0.387 & 37.96 & 0.665 & 92.03 & \textbf{0.059}& 5.08 & \textbf{0.028} & 4.97\\
MAT ~\cite{li2022mat} & 0.331 & 32.55 & 0.280 & 20.63 & 0.479 & 82.37 & 0.080 & 5.16 & 0.077 & 4.95\\
GLaMa~\cite{lu2022glama} & 0.327 & 30.76 & 0.289 & 18.61& 0.481 & 80.44 & 0.081 & 5.83  & 0.080 & 5.10\\
FcF~\cite{jain2023keys} & 0.305 & 27.95 & 0.378 & 31.91 & 0.502 & 73.24 &0.086 & 4.63 & 0.071 & 4.42\\
DDIM+CS (\textbf{ours}) & \textbf{0.272} & \textbf{20.37} & \textbf{0.259} & \textbf{15.33}& \textbf{0.372} & \textbf{39.05} & 0.079 & \textbf{4.21} & 0.064 & \textbf{3.58}\\
\hline
\end{tabular}
}
\caption{\textbf{Results of face inpainting} on CelebA-HQ-256. DDIM with our compensation sampling shows consistent improvement over state-of-the-art methods, for both LPIPS and FID-50k metrics. Best results in \textbf{bold}. \vspace{-2mm}}
\label{tab:4}
\end{table*}

\textbf{Qualitative evaluation}. Images generated by DDIM with compensation sampling for CelebA-64 and FFHQ-256 appear in Figure~\ref{fig:2} (left). Images generated for CIFAR-10 are shown in the supplementary material. For different resolutions, our method generates images with realistic details. We show the stochastic process with a wide diversity in outputs in Figure~\ref{fig:2} (right). We present more examples, including failure cases, in the supplementary material.

\begin{figure}[htb]
\centering
\includegraphics[width=\columnwidth]{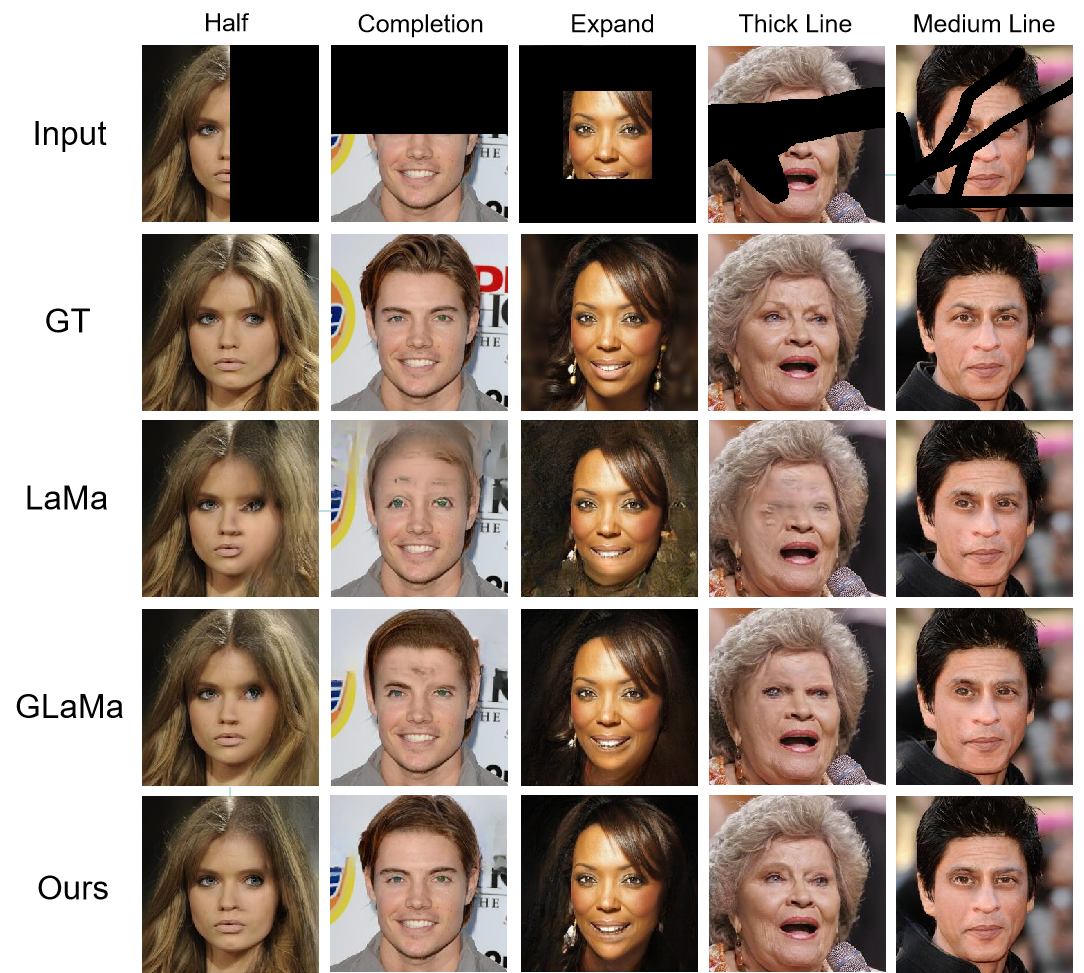}
\caption{\textbf{Visualization of inpainting results on CelebA-HQ-256.} Compared to state-of-the-art methods, our method generates more realistic images in different mask situations.}
\label{fig:3}
\end{figure}

\subsection{Face inpainting}
The goal of face inpainting is to restore missing or damaged parts in a face image, resulting in a complete facial image.

\textbf{Experiment setting}. We perform the inpainting experiment with DDIM+CS on CelebA-HQ-256 with NFE = 100 and 40 training epochs for the compensation module to reduce the output diversity. Results with different numbers of epochs for the compensation module appear in the supplementary material. All other training hyper-parameters are the same as for the unconditional generation experiment. We employ the same schedule of corruption transforms as in previous works~\cite{batzolis2021conditional,lugmayr2022repaint}. Each training image of the reconstruction model is corrupted with a synthetically generated mask. The superimposition process starts with input images $x_0$, which are iteratively masked for $T$ steps via multiplication with a sequence of masks $z_{\alpha_i}$ with increasing $\alpha_i$. We follow~\cite{lu2022glama,lugmayr2022repaint} and use five mask types: \textit{Half}, \textit{Completion}, \textit{Expand}, \textit{Thick Line}, and \textit{Medium Line}, see Figure~\ref{fig:3}. 

\textbf{Evaluation metrics}. Following recent image inpainting literature, we use Learned Perceptual Image Patch Similarity (LPIPS)~\cite{zhang2018unreasonable} and FID as similarity metrics. Compared to PSNR and SSIM~\cite{wang2004image}, LPIPS and FID are more suited to  measure the performance of inpainting for large masks~\cite{lu2022glama}.

\textbf{Quantitative evaluation.} Results of diffusion and GAN-based approaches appear in Table~\ref{tab:4}. We report results from the papers, and use open source implementations to calculate missing numbers. Our DDIM+CS outperforms other approaches in almost all cases. 
Although we achieve slightly worse LPIPS scores than RePaint for the Thick Line and Medium Line masks (0.059$\rightarrow$0.079, 0.028$\rightarrow$0.064), we show better performance for other masks. Moreover, we consistently outperform all tested methods on FID-50k score. Models trained with Expand masks give worse results than other mask types, which is consistent with observations in GLaMa~\cite{lu2022glama}. Still, our approach significantly reduces both LPIPS and FID scores.

\textbf{Qualitative evaluation}. Inpainting results appear in Figure~\ref{fig:3} and the supplementary material. Compared to state-of-the-art approaches LaMa~\cite{suvorov2022resolution} and GLaMa~~\cite{lu2022glama}, our generated images are more realistic and detailed.

\subsection{Face de-occlusion}
\begin{figure}[htb]
\centering
\includegraphics[width=\columnwidth]{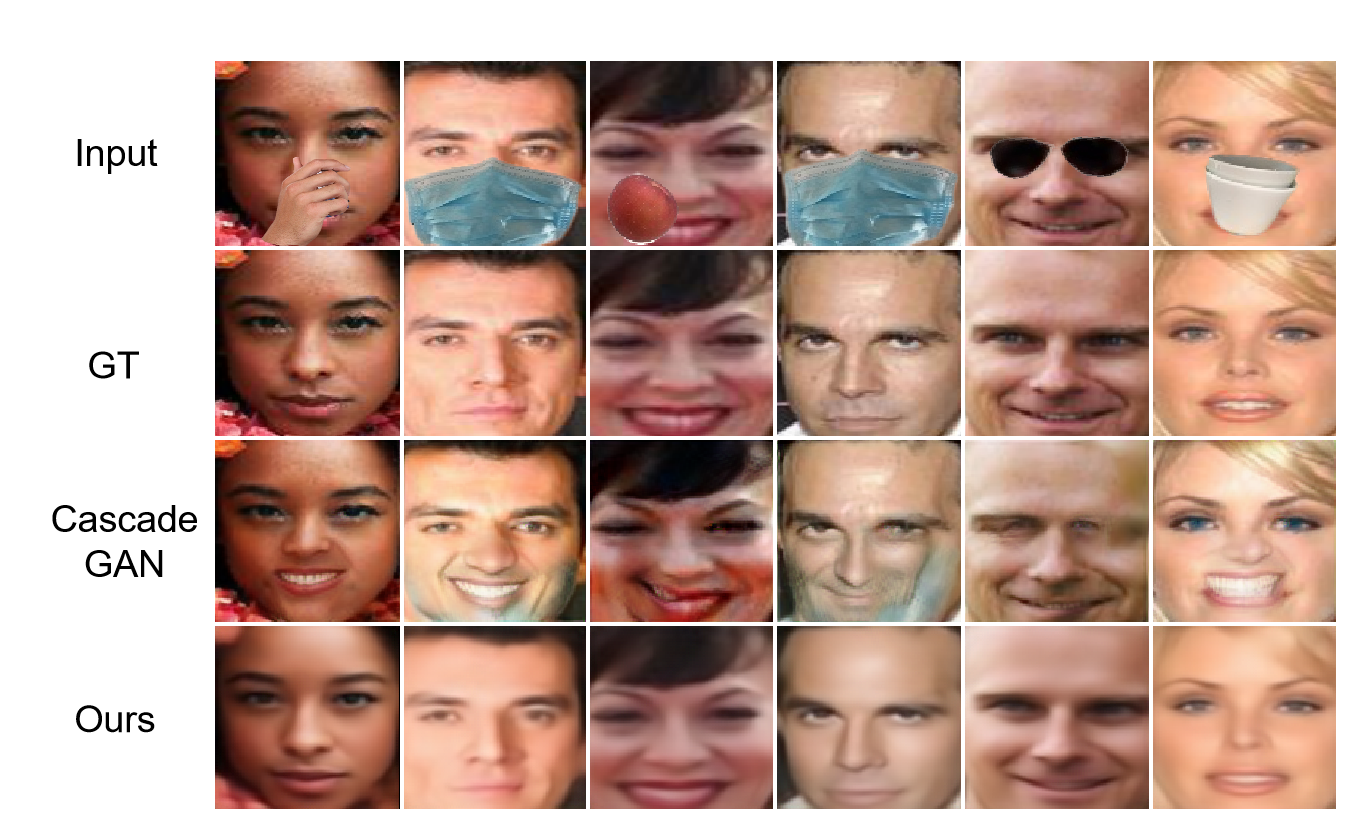}
\caption{\textbf{Face de-occlusion results on FSG}. Our results are closer to the ground truth than top-performing Cascade GAN. \vspace{-2mm}}
\label{fig:4}
\end{figure}

We further experiment with face de-occlusion. Compared to face inpainting, face de-occlusion is more challenging since the missing contents of the face images are not black pixels but pixels with values similar to faces, as shown in Figure~\ref{fig:4}. During the reconstruction process, the models therefore not only have to infer which parts belong to the face, but also fill in the content of the missing parts. 

\textbf{Experiment setting}. We train and evaluate our DDIM+CS using FSG (200k images)~\cite{cheung2021facial}. The face images are synthesized with common occlusion objects that are semantically placed relative to facial landmarks. We use the same setting as in the face inpainting experiments.

\textbf{Evaluation metrics}. While face de-occlusion has been addressed with GANs, we believe ours is the first work to apply diffusion models to face de-occlusion without any additional information such as a segmentation mask. To compare to other works, we use PSNR and SSIM as metrics.

\begin{table} [htb]
\centering
\resizebox{0.8\linewidth}{!}{
\begin{tabular}{|l|c|c|}
\hline
 \textbf{Method} & \textbf{PSNR} $\uparrow$ & \textbf{SSIM} $\uparrow$\\
\hline
Occluded image & 10.4764 & 0.6425\\
CycleGAN~\cite{zhu2017unpaired} & 13.7667 & 0.6459 \\
DeepFill~\cite{yu2018generative} & 14.5140 & 0.7029 \\
OA-GAN~\cite{dong2020occlusion} & 17.1828 & 0.7215 \\
FSG-GAN~\cite{cheung2021facial} & 21.1112 & 0.7936 \\
Cascade GAN~\cite{zhang2022face} & 26.4736 & 0.8422\\
SRNet~\cite{yin2023segmentation} & 27.0031 & 0.8493\\
\hline
DDIM+CS (\textbf{ours}) & \textbf{31.3842} & \textbf{0.8699}\\
\hline
\end{tabular}
}
\caption{\textbf{Face de-occlusion on FSG}. Our method achieves consistent improvement over GANs. Best results in \textbf{bold}.}
\label{tab:5}
\end{table}

\textbf{Quantitative evaluation}. We summarize our and state-of-the-art GAN results in Table~\ref{tab:5}. DDIM with compensation sampling shows significantly better performance than any of the tested GANs. For example, DDIM+CS achieves ~20\% lower PSNR compared to the best tested SRNet.

\textbf{Qualitative evaluation}. The results of DDIM+CS and Cascade GAN are shown in Figure~\ref{fig:4}. Our generated images are realistic and occluding objects are generally fully removed. However, our results are more blurred, which could be caused by our use of the naive loss function in Equation~\ref{equa:3}, while 
other works like Cascade GAN~\cite{zhang2022face} employ sophisticated hybrid loss functions including pixel-wise face loss~\cite{yin2023segmentation}, background loss~\cite{yin2023segmentation}, identity loss~\cite{yin2023segmentation}, SSIM loss~\cite{wang2004image}, and total variation loss~\cite{mahendran2015understanding}.

\begin{figure}[htb]
\centering
\includegraphics[width=\columnwidth]{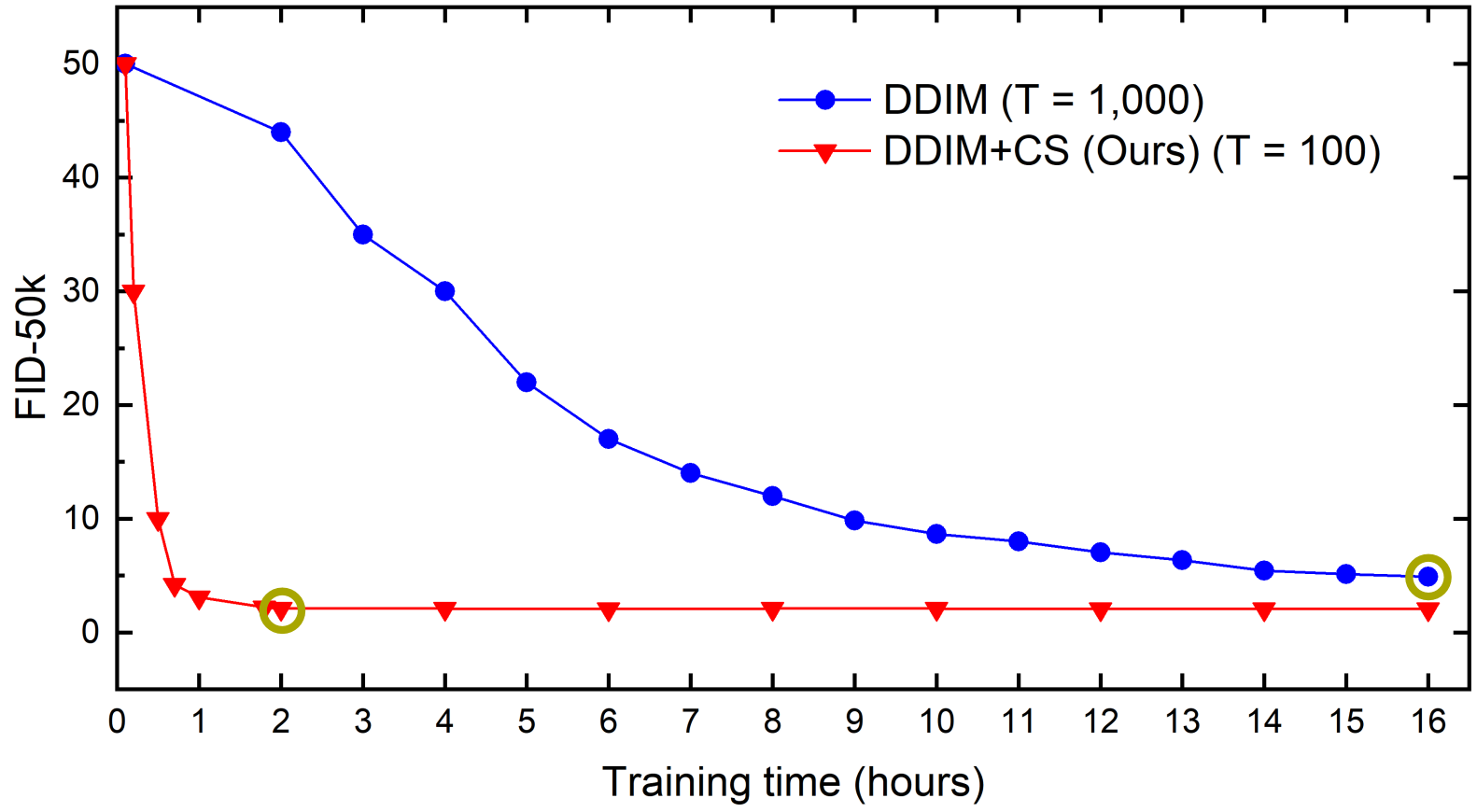}
\caption{\textbf{Training time comparison with DDIM} for unconditional face generation on CelebA-64. DDIM and DDIM+CS are trained with 1,000 and 100 time steps, respectively.}
\label{fig:5}
\end{figure}

\subsection{Training computation cost} \label{sec:exp_computation_cost}
We consistently observe competitive or better results when using compensation sampling with 10 times fewer time steps. Here, we investigate the benefit of our algorithm in accelerating the training process. Figure~\ref{fig:5} clearly illustrates the value of our contribution. It shows FID-50k scores during the DDIM training process with common and compensation sampling for unconditional generation on CelebA-64. DDIM+CS ($T=100$) takes only two hours to converge to a FID-50k of 2.11, while DDIM ($T=1,000$) takes 16 hours to achieve a score of 4.88.

\begin{table} [htb]
\centering
\resizebox{1.0\linewidth}{!}{
\begin{tabular}{|c|l|c|c|c|}
\hline
Method & Property & CIFAR-10 & CelebA-64 & FFHQ-256 \\
\hline
\multirow{3}{*}{\parbox{2cm}{\centering DDIM\quad ($T=1,000$)}} & GFLOPS & 7.76 & 15.52 & 248.17 \\
& Total time & 7h & 16h & 26h \\
& FID-50k & 3.95 & 4.88 & 8.41 \\
\hline
\multirow{3}{*}{\parbox{2cm}{\centering DDIM+CS\quad(\textbf{ours})\quad ($T=100$)}} & GFLOPS & 7.78 & 15.58 & 249.07 \\
& Total time & 0.4h & 2.0h & 5.3h \\
& FID-50k & 2.01 & 2.11 & 4.02 \\
\hline
\end{tabular}
}
\caption{\textbf{Training time} for DDIM and ours on different datasets/resolutions. Four NVIDIA Tesla A100 GPUs were used.}
\label{tab:computation cost}
\end{table}

In Table~\ref{tab:computation cost}, we report the training times and FLOPs for three datasets with different resolutions. Using compensation sampling is consistently much faster as a result of the fewer time steps that are required. The limited increase in FLOPs demonstrates that the training of the compensation module has a negligible effect on the computation cost.

\subsection{Ablation study}
\begin{figure}[htb]
\centering
\includegraphics[width=0.45\textwidth]{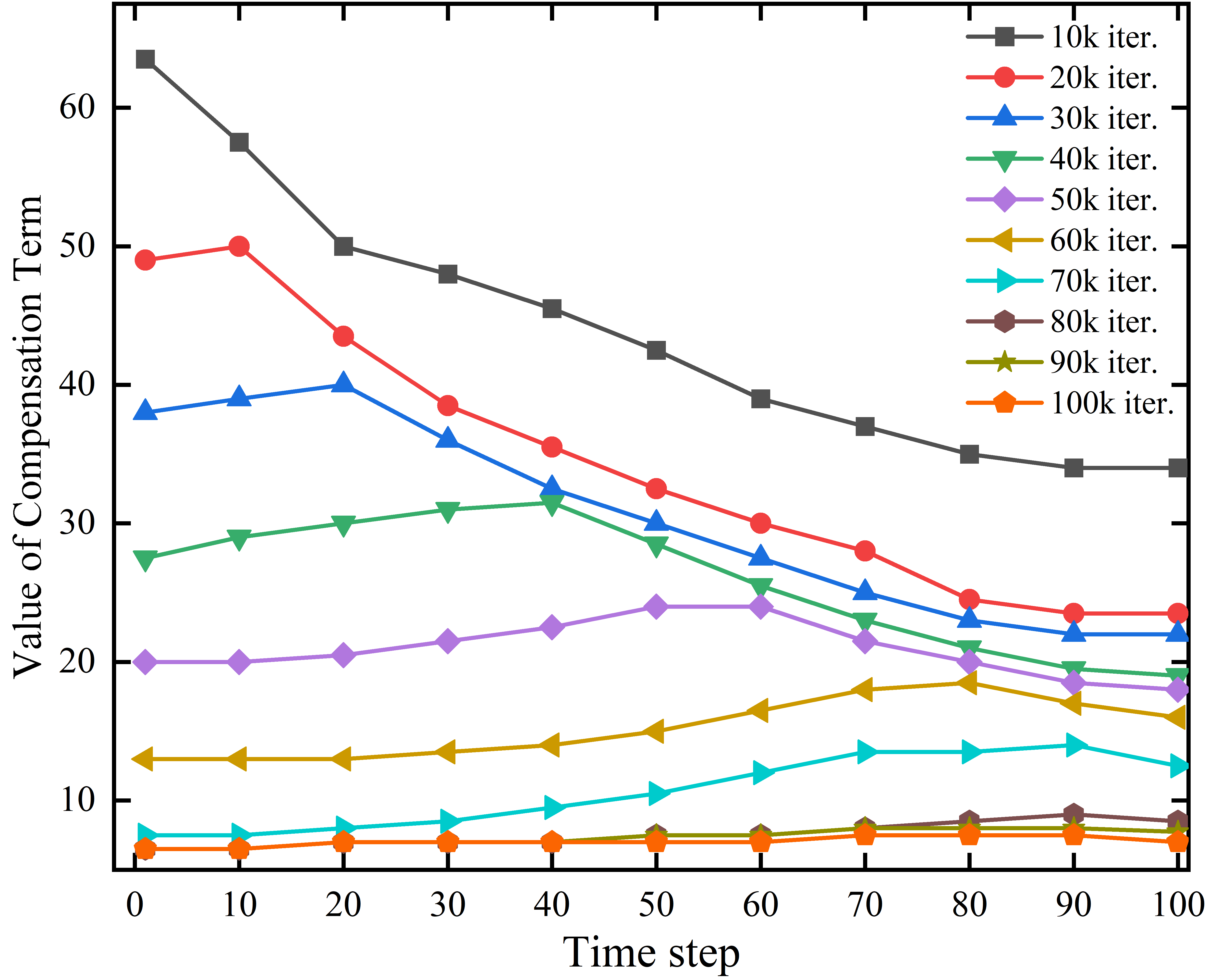}
\caption{\textbf{Compensation term value per training iteration}, unconditional face generation on FFHQ-256, DDIM+CS $T=100$. \vspace{-3mm}}
\label{fig:7}
\end{figure}

\textbf{Compensation term value during training}. We study how the compensation term varies during DDIM+CS training for unconditional generation on FFHQ-256. In Figure~\ref{fig:7}, we show the average compensation term value over all training images every 10 time steps, for a total of $T = 100$ steps.

We make the following conclusions. First, the value of the compensation term continuously decreases at each time step because the accumulated error is higher in early training, and more compensation is needed. Second, the highest value gradually moves from time step $0$ to $100$ as training progresses. The compensation is initially aimed at recovering the fine details, before focusing on the more difficult task of generating structure, which is also reported in~\cite{choi2022perception}.

\textbf{Role of compensation term}. We evaluate the role of the compensation term during training and generation, on unconditional generation on FFHQ-256, $T=100$. We analyze common sampling, which corresponds to DDIM~\cite{song2020score}. We also compare to Cold Diffusion~\cite{bansal2022cold}. For DDIM with compensation sampling, we investigate the setting in which the compensation term is used during training but not generation (DDIM+CS (T)), and the default setting in which it is used in both (DDIM+CS).

From Table~\ref{tab:ablation_sampling}, we observe a significant increase in performance when using compensation sampling. This difference is mainly due its use during training. When also using the compensation term in the generation phase, the FID-50k score slightly improves (11.94$\rightarrow 11.89$). The worse performance of Cold Diffusion might be caused by incorrect assumptions about the noise distribution. These results demonstrate that the compensation term is beneficial in producing higher-quality outputs. However, the added value of the compensation term in inference is minimal. This makes sense, since its value decreases as the training progresses, see Figure~\ref{fig:5}. Using a small residual term in the sampling during generation therefore has a limited effect.

\begin{table} [htb]
\centering
\resizebox{1.0\linewidth}{!}{
\begin{tabular}{|c|l|c|c|c|}
\hline
& DDIM & Cold Diffusion & DDIM+CS (T) & DDIM+CS \\
\hline
FID-50k & 17.89 & 31.32 & 11.94 & 11.89 \\
\hline
\end{tabular}
}
\caption{\textbf{Role of compensation term in unconditional generation on FFHQ-256}. All methods use $T=100$ time steps.}
\label{tab:ablation_sampling}
\end{table}

\section{Conclusions} \label{sec:conclusions}
We have introduced compensation sampling, a novel algorithm to guide the training of diffusion models. Our innovation has three main benefits. First, models with compensation diffusion converge to a better solution because we continuously reduce the accumulated error during training. This is demonstrated by higher-quality generated images. Second, by guiding the convergence, we can reduce the number of time steps up to an order of magnitude. Finally, our approach can be applied broadly in generation tasks. We have evaluated our method on unconditional face generation, face inpainting, and face de-occlusion. Our results on benchmark datasets consistently demonstrate superior performance and increased efficiency compared to state-of-the-art diffusion and GAN models. Our sampling approach is general and can be used in a wide range of diffusion models.

\small

\input{main.bbl}
\bibliographystyle{ieeenat_fullname}

\clearpage

\pagenumbering{gobble} 
\setcounter{page}{1} 

\input{supplementary}

\end{document}

%% file: preamble.tex
\usepackage[dvipsnames]{xcolor}


%% file: supplementary.tex
\clearpage
\setcounter{page}{1}
\maketitlesupplementary

\setcounter{section}{0} 
\renewcommand{\thesection}{\arabic{section}} 

\section{Overview}
\label{sec:rationale}
To demonstrate the merits of our compensation sampling approach beyond tasks related to face reconstruction, we present additional experiments on unconditional general image generation using CIFAR-10. We then present additional quantitative and qualitative results for the three tasks in the main paper: unconditional face generation, face inpainting, and face de-occlusion. We provide more analysis of the inner workings of our approach, including a comparison of the computation cost, the effect of the number of training epochs of the compensation module, and a brief summary of the model architecture.

\section{Unconditional generation on CIFAR-10} \label{sec:app_cifar}
\subsection{Experimental setting}
To evaluate on a non-face generation task, we use the more general CIFAR-10 (60k images)~\cite{krizhevsky2009cifar10} dataset. For fair comparison, our training hyper-parameters including batch size and decay rate are the same as in DDPM~\cite{ho2020denoising}. We again use Gaussian noise as our corruption mechanism, and adopt the fixed linear variance schedule $\beta_1, ..., \beta_T$ as in DDPM for the prior noising process. In line with the experiments on unconditional face generation, we reduce the training time steps for our model to $T=100$. Both the compensation module and diffusion model use the Adam optimizer. Experiments are conducted on 4 NVIDIA Tesla A100 GPUs.

\begin{figure}[htb]
\centering
\includegraphics[width=1.0\columnwidth]{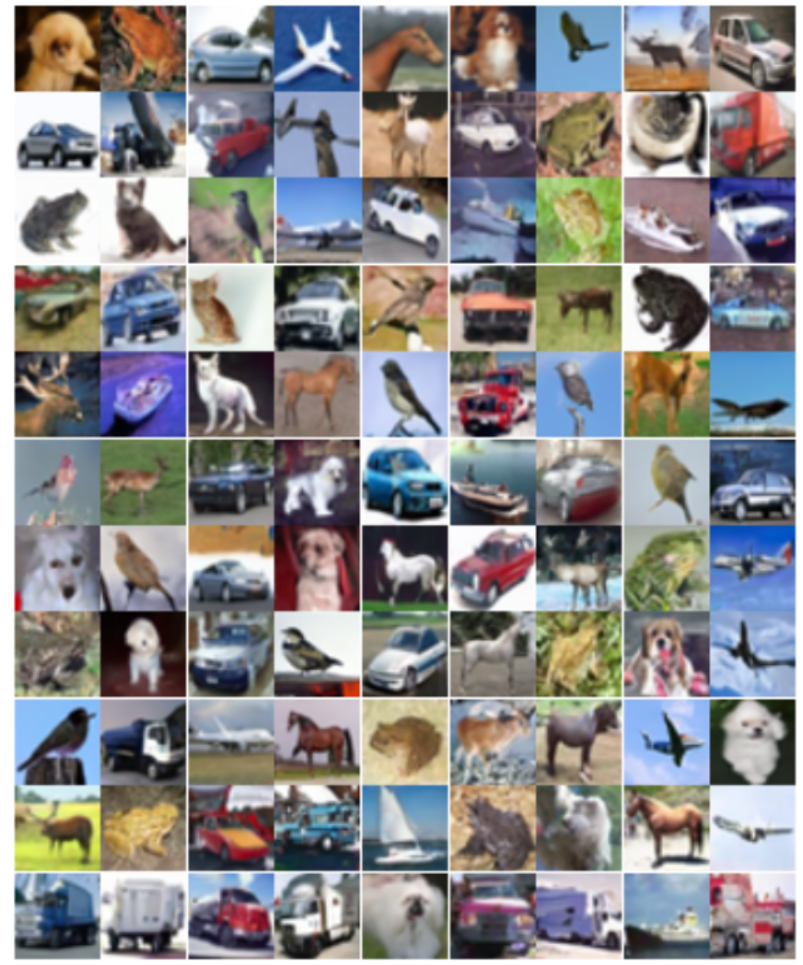}
\caption{\textbf{Generated images from our model trained on CIFAR-10}. Note the diversity of the various classes.}
\label{fig:1}
\end{figure}

\subsection{Qualitative results}
We show generated images of our method for CIFAR-10 in Figure~\ref{fig:1}. We observe a wide diversity of our method, and the images generally have good realism and quality.

\begin{figure}[t]
\centering
\includegraphics[width=1.0\columnwidth]{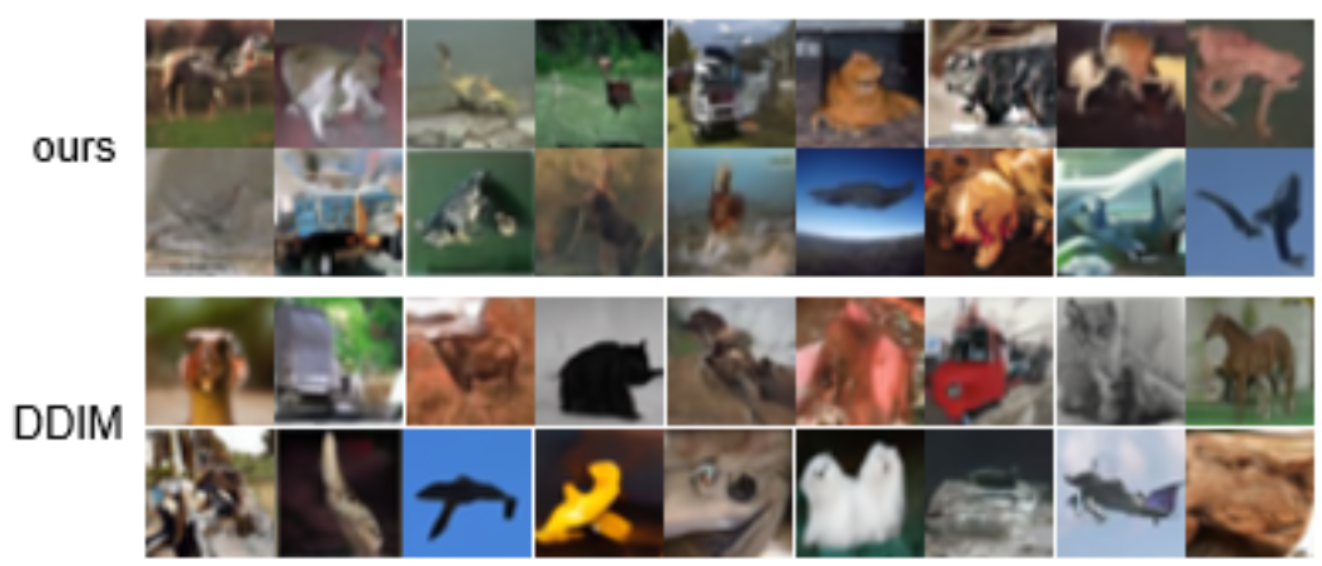}
\caption{\textbf{Failure cases of DDIM and ours on CIFAR-10}.}
\label{fig:2}
\end{figure}

\begin{figure}[t]
\centering
\includegraphics[width=\columnwidth]{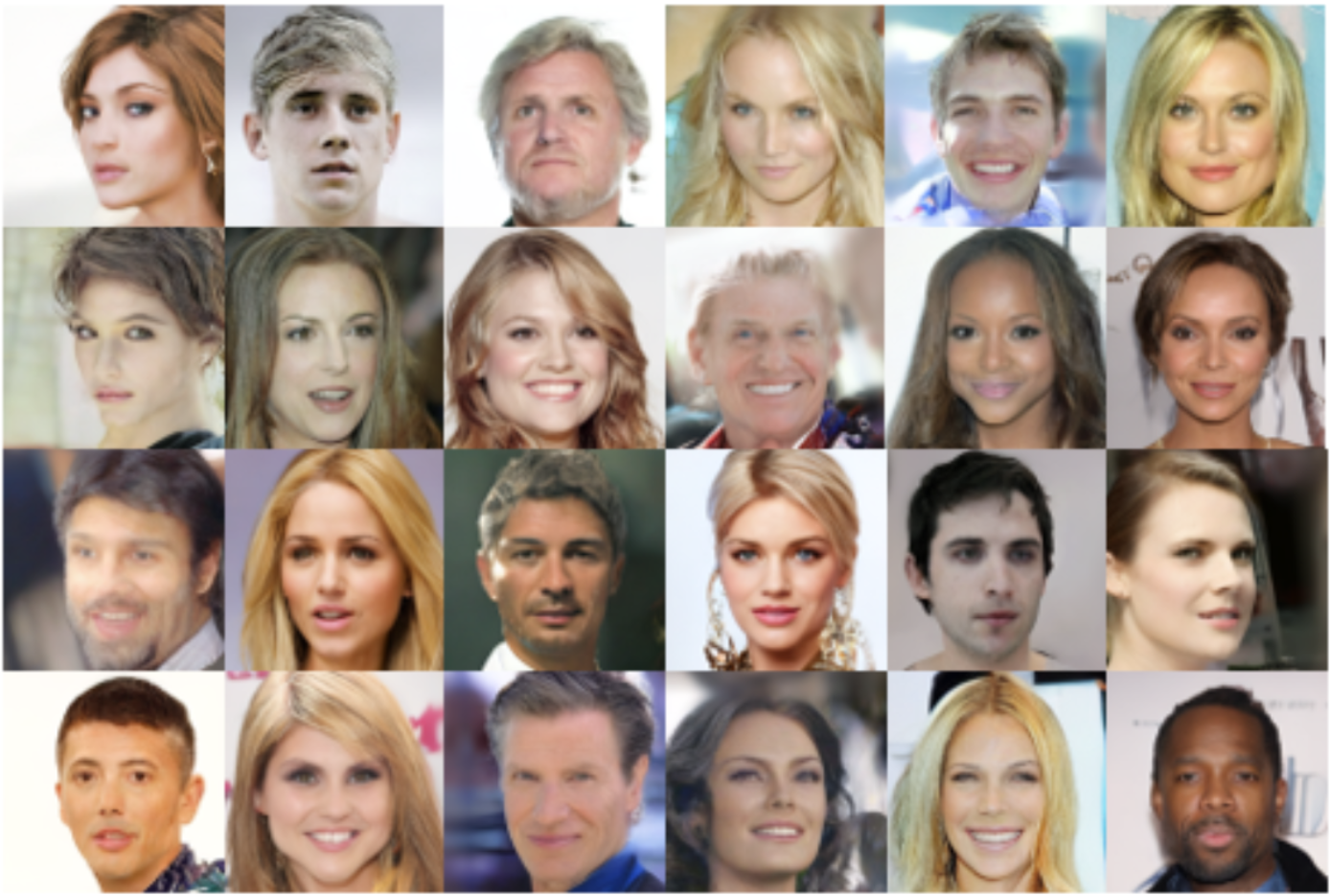}
\caption{\textbf{Unconditional face generation on CelebA-64}. More samples with compensation sampling.}
\label{fig:3}
\end{figure}

\begin{figure}[t]
\centering
\includegraphics[width=\columnwidth]{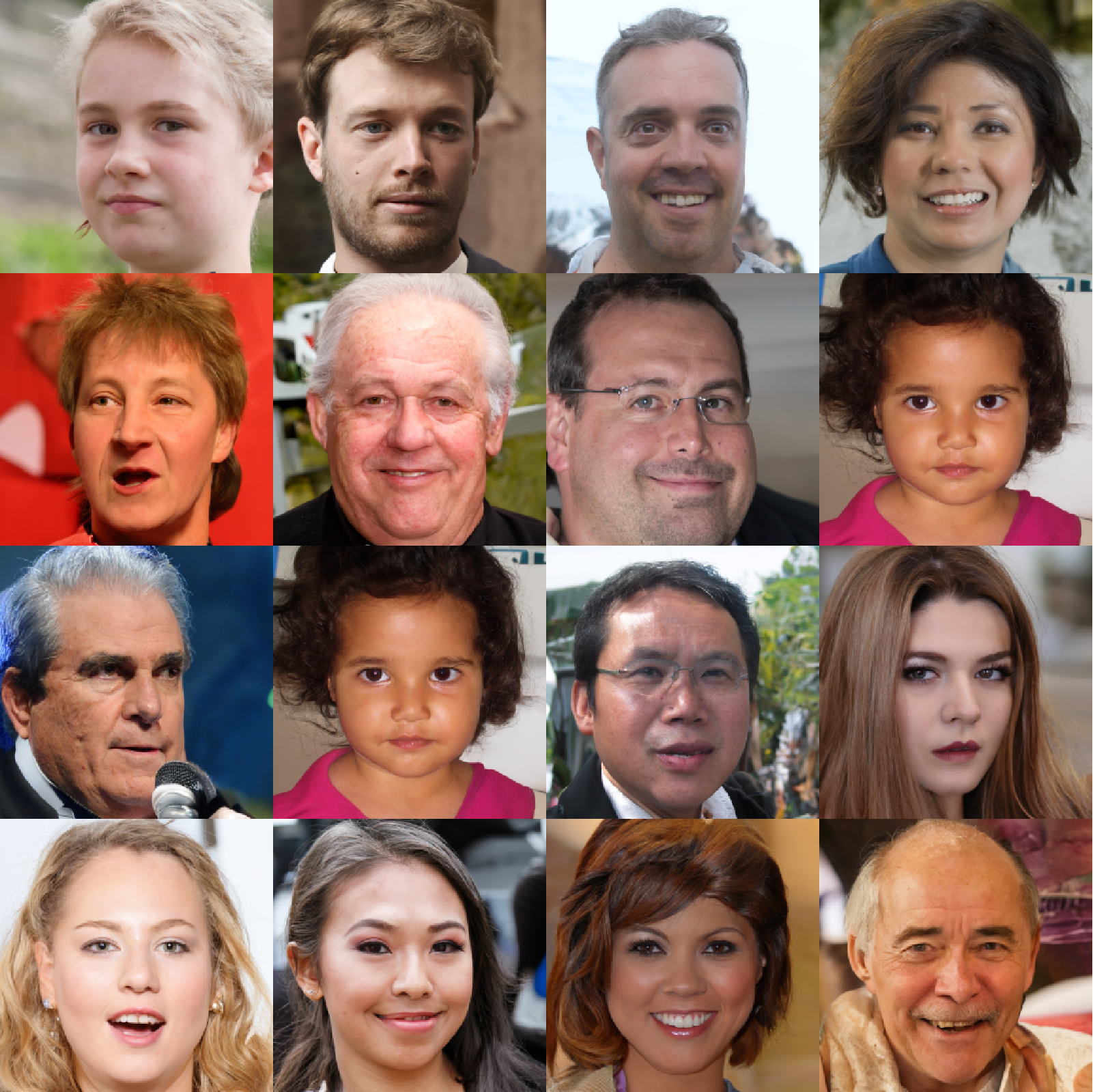}
\caption{\textbf{Unconditional face generation on FFHQ-256}. More samples with compensation sampling.
}
\label{fig:4}
\end{figure}

\subsection{Failure cases}
We show and compare failure cases of DDIM and ours on CIFAR-10 in Figure~\ref{fig:2} to better understand the limitations of either model. 

Both DDIM and our model can generate images that are hard to classify. Specifically, DDIM tends to generate unknown animals, while our method tends to generate images of animals that combine features from different animals. For example, the first picture in the first image seems a combination of a horse and a deer. However, considering that the network has no additional classification information, conditioning the generation on the class may largely prevent this issue. For both models, we observe that images of classes without articulation, such as cars and planes, are typically more realistically generated.

\section{Unconditional face generation} \label{sec:app_unconditional}
\subsection{Additional results}
\begin{figure}[t]
\centering
\includegraphics[width=\columnwidth]{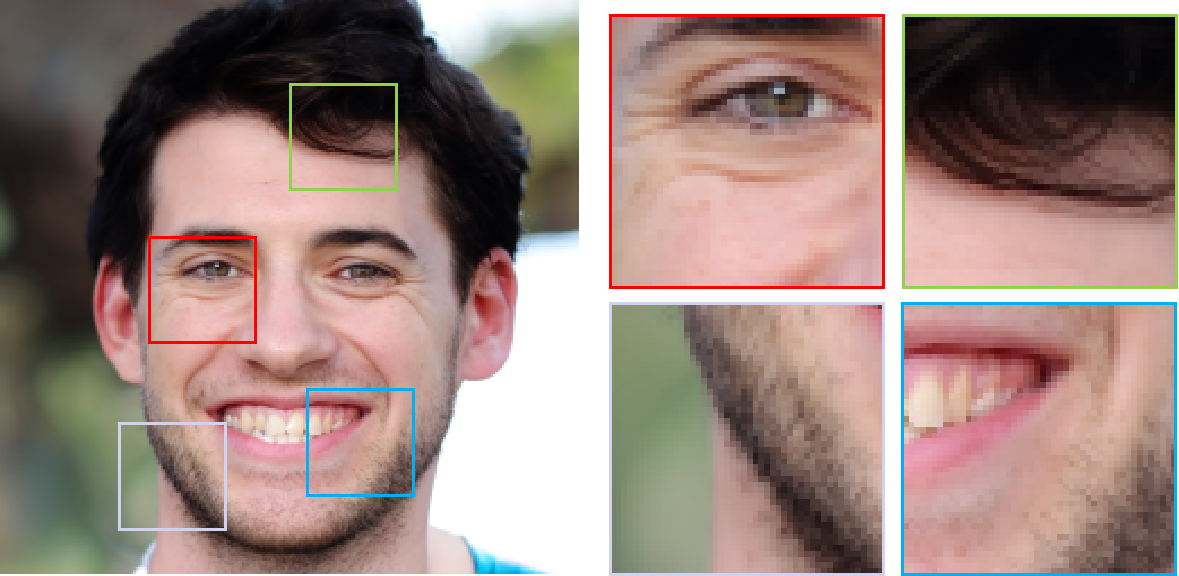}
\caption{\textbf{Close-up of sample face} generated after training on FFHQ-256. Note the realistic details in all parts of the face.}
\label{fig:4-1}
\end{figure}

We show more generation results on CelebA-64 and FFHQ-256 in Figures~\ref{fig:3} and \ref{fig:4} with the same setting in the main paper. Again, the diversity of the faces becomes apparent. With a close-up in Figure~\ref{fig:4-1}, we can see the high quality details of the generated image. In particular, we observe realistic facial hair and wrinkles around the eyes and mouth.

\begin{figure}[t]
\centering
\includegraphics[width=\columnwidth]{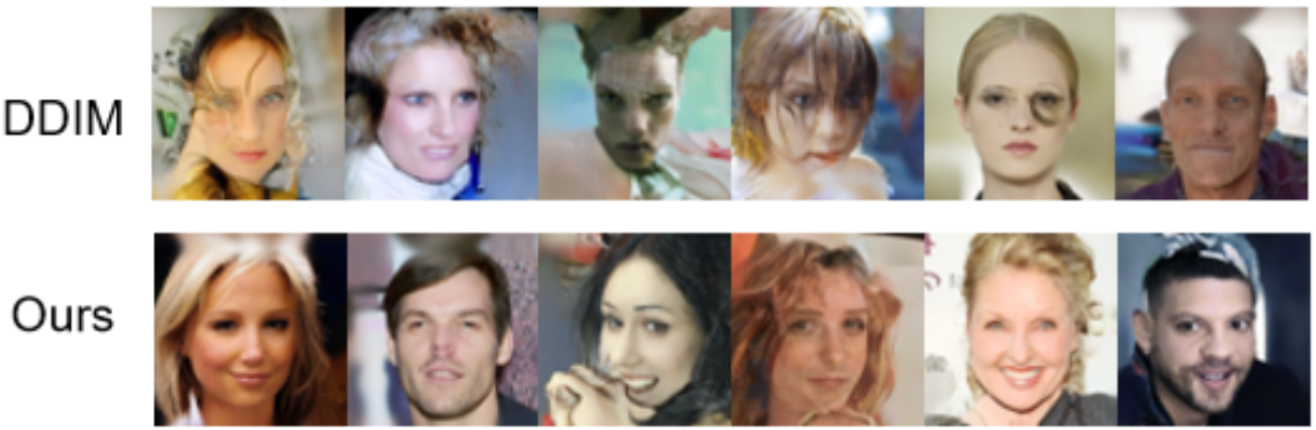}
\caption{\textbf{Failure cases of DDIM and our model on CelebA-64}.}
\label{fig:5}
\end{figure}

\begin{figure}[t]
\centering
\includegraphics[width=\columnwidth]{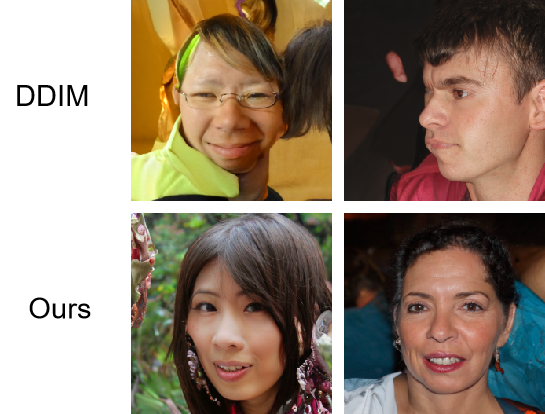}
\caption{\textbf{Failure cases of DDIM and our model on FFHQ-256}.}
\label{fig:5-1}
\end{figure}

\subsection{Failure cases}
We also present some failure cases of DDIM and our model in Figures~\ref{fig:5} and \ref{fig:5-1} to better analyze the patterns. Similar to DDIM, our method's failures mainly focus on the details of face or hands. In addition, DDIM tends to have structural errors, where the face shapes are corrupted. For example, the first face of DDIM in Figure~\ref{fig:5} is incomplete, and the second face of DDIM in Figure~\ref{fig:5-1} is missing an eye.

\section{Face inpainting on CelebA-HQ} \label{sec:app_inpainting}
\begin{figure}[t]
\centering
\includegraphics[width=\columnwidth]{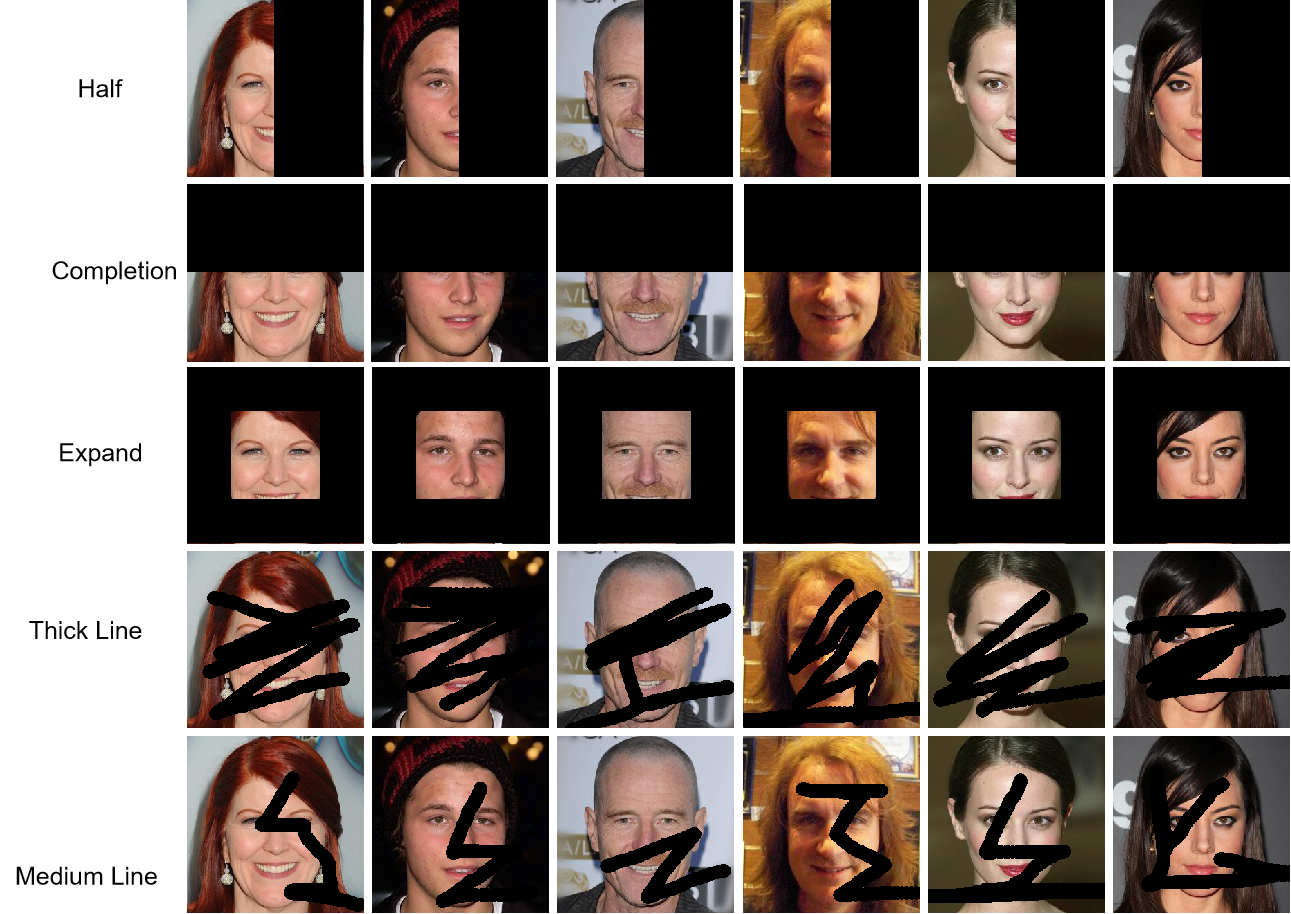}
\caption{\textbf{Inputs for face inpainting experiment}. Each column represents a different person.}
\label{fig:6-1}
\end{figure}

\begin{figure}[htbp]
\centering
\includegraphics[width=\columnwidth]{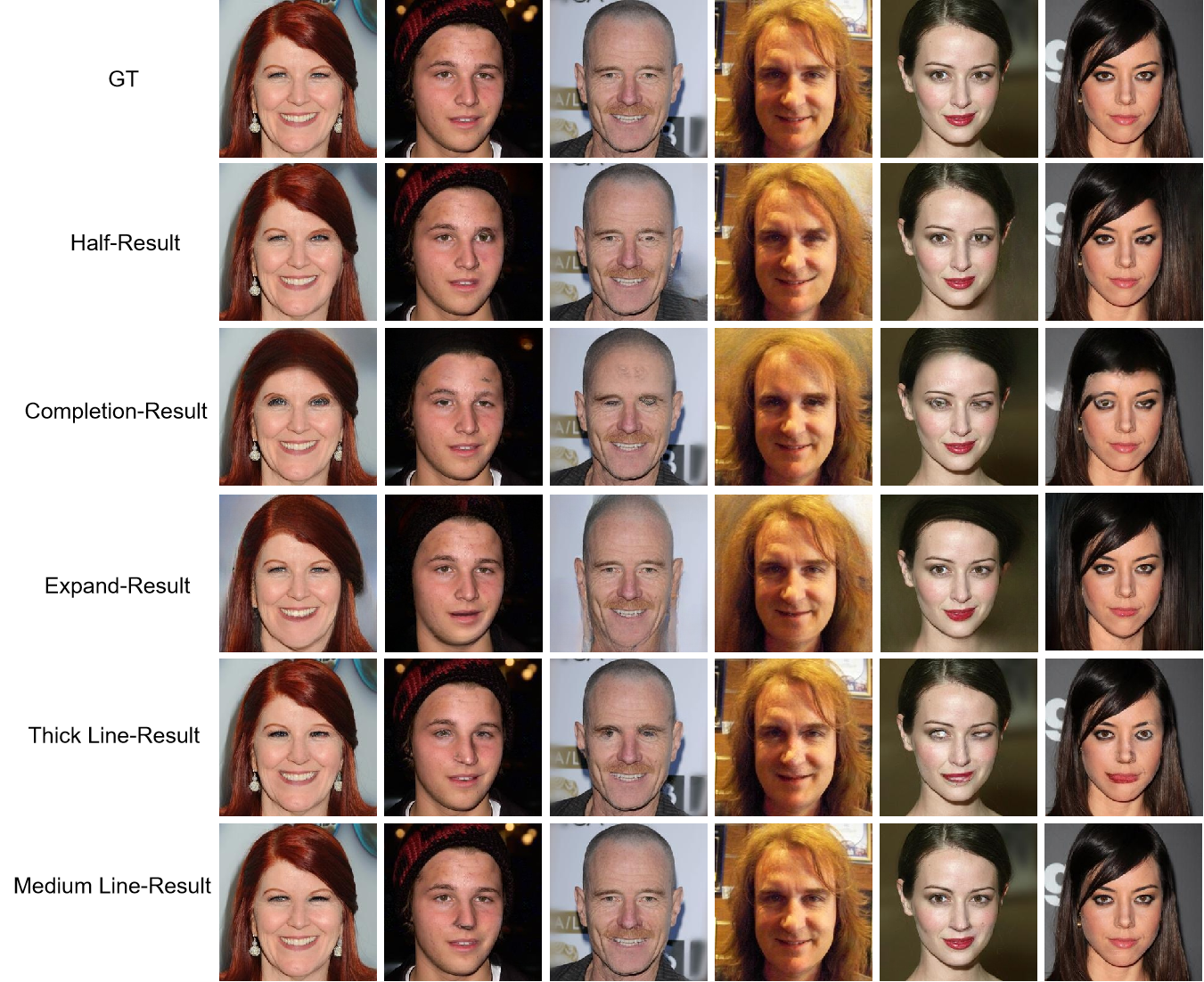}
\caption{\textbf{Results on face inpainting} obtained with our model trained on CelebA-HQ-256. Each column represents a different person.}
\label{fig:6-2}
\end{figure}

We show more face inpainting results on CelebA-HQ-256 in Figures~\ref{fig:6-1} and \ref{fig:6-2}, where Figure~\ref{fig:6-1} shows the input and the ground truth, and Figure~\ref{fig:6-2} shows the generation results. Similar to the previous face inpainting task, we use more epochs to train the compensation module to generate pixels close to human faces, which can reduce the diversity of results to some extent. The failures mainly happen when large areas are occluded. In some cases, the eyes are not realistically recovered.

\section{Face de-occlusion on FSG} \label{sec:app_de-occlusion}
\begin{figure}[htb]
\centering
\includegraphics[width=\columnwidth]{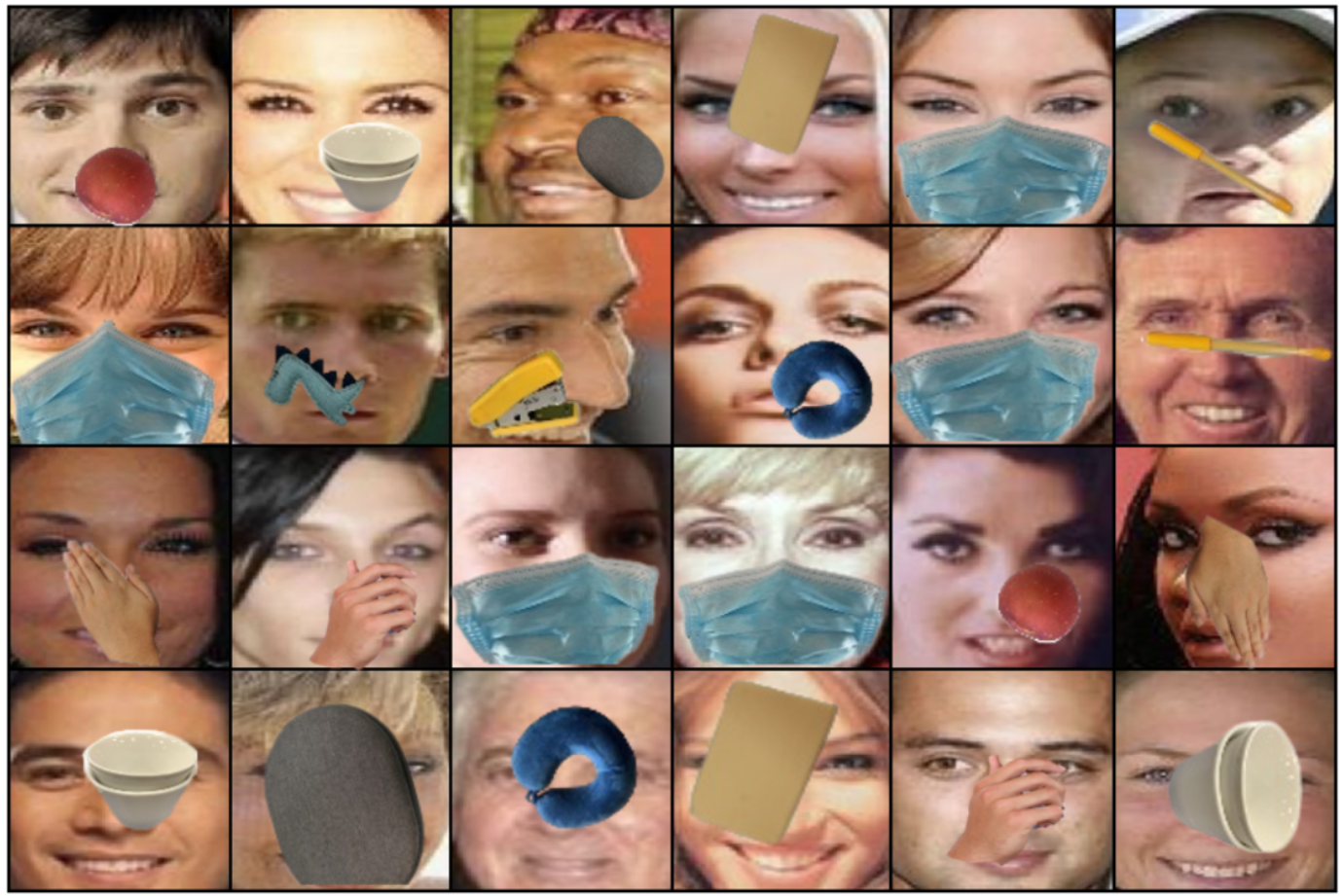}
\caption{\textbf{Inputs for face de-occlusion experiment} from FSG.}
\label{fig:8-1}
\end{figure}

\begin{figure}[htb]
\centering
\includegraphics[width=\columnwidth]{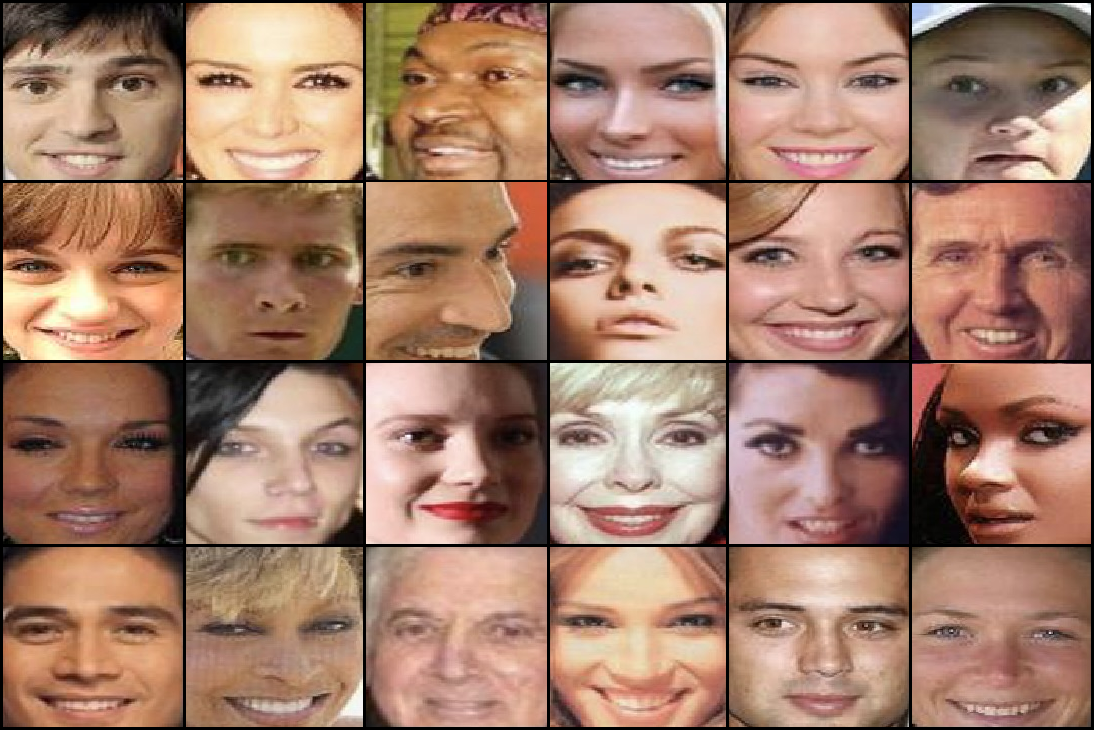}
\caption{\textbf{Ground truth for face inpainting} obtained with our model trained on FSG.}
\label{fig:8-2}
\end{figure}

\begin{figure}[htb]
\centering
\includegraphics[width=\columnwidth]{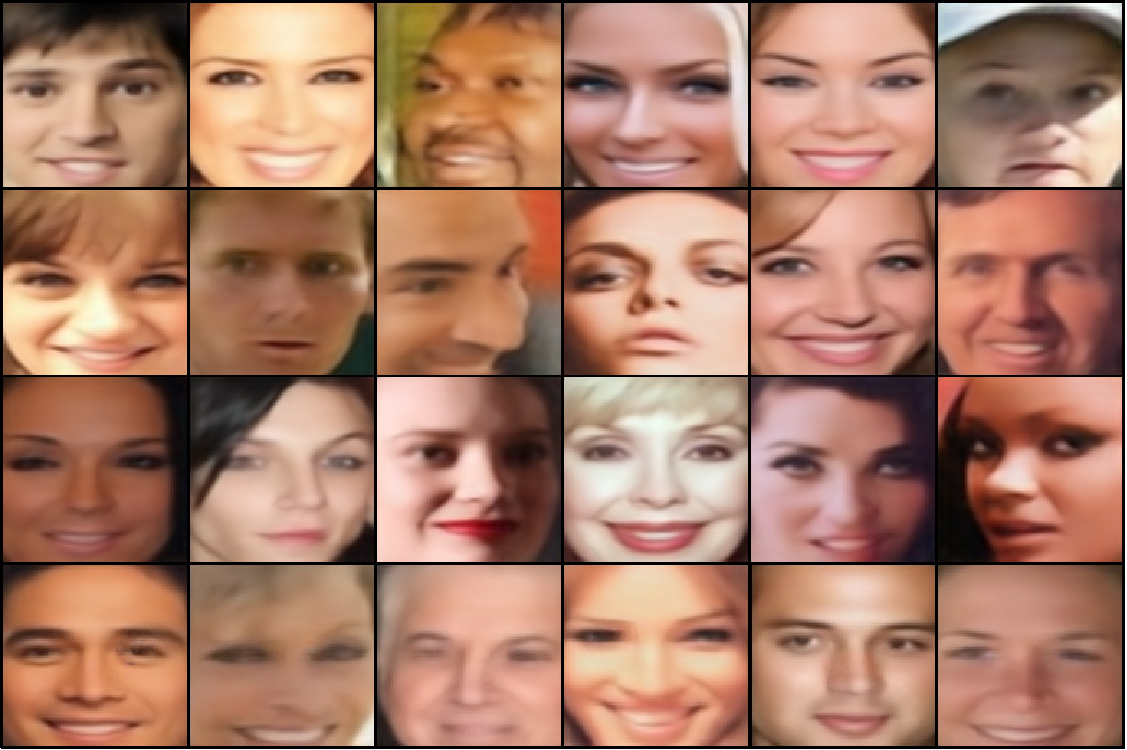}
\caption{\textbf{Results on face inpainting} obtained with our model trained on FSG.}
\label{fig:8-3}
\end{figure}

Similar to the face inpainting task, we show more face de-occlusion results on the FSG dataset with the same setting in Figures~\ref{fig:8-1}, \ref{fig:8-2}, and \ref{fig:8-3}. Figure~\ref{fig:8-1} shows the input, Figure~\ref{fig:8-2} the ground truth, and Figure~\ref{fig:8-3} shows the results generated with our model. Our results seem blurred, and failure cases occur especially when the top face area is occluded.

\section{Comparison of computation cost} \label{sec:app_computation_cost}
Although our method can highly accelerate the training process and improve the image quality because fewer time steps are required and the accumulated reconstruction error is reduced, we also introduce an additional compensation module into our method during the training process. This increases the computation cost. To assess the computation cost during the training process, we conduct the unconditional generation task on various datasets on 4 NVIDIA Tesla A100 GPUs, and compare the FLOPs and time consumption of our method and DDIM during the training process, and summarize the result in Table~\ref{tab:2}.

From the table, it can be observed that the introduction of the compensation module barely increases the FLOPs of our models, which validates that the compensation module has a negligible effect on the computation cost. Furthermore, our total time consumption for the training process is much lower than the DDIM (26 hours$\rightarrow$5.3 hours on FFHQ-256, 19 hours$\rightarrow$3.9 hours on FSG), which means that our approach takes much less training time than DDIM to reach the convergence.

\begin{table}[htb]
\centering
\resizebox{1.0\linewidth}{!}{
\begin{tabular}{|c|c|c|c|c|c|c|}
\hline
\multirow{2}{*}{Iter.} & \multicolumn{3}{c|}{CelebA-64} & \multicolumn{3}{c|}{FFHQ-256} \\
  & FID & Pre. & Recall & FID & Pre. & Recall \\
\hline
1 & 2.12&0.79&0.71     &11.89&0.76&0.65      \\
2& 2.13&0.79&0.69     &11.89&0.76&0.64      \\
5& 2.15&0.82&0.52     &11.93&0.78&0.55      \\
10& 2.15&0.83&0.48     &11.87&0.78&0.50      \\
20& 2.13&0.84&0.44    &11.93&0.80&0.45      \\
40& 2.12&0.85&0.41     &11.92&0.80&0.43     \\
80& 2.12&0.85&0.40     &11.89&0.79&0.43     \\
\hline
\end{tabular}
}
\caption{\textbf{Effect of training epochs of the compensation module} on unconditional and conditional generation.}
\label{tab:6}
\end{table}

\section{Effect of training epochs of compensation module}\label{sec:app_training_epochs}
In the main paper, we use a single training epoch for the compensation module in the unconditional generation experiment, so that we can introduce additional noise to increase the diversity of the generation. In the face inpainting and face de-occlusion tasks, we use more training epochs for the compensation module, so the compensation term will generate pixels close to human faces, thus we have a higher chance to get realistic human faces.

Considering the compensation term will encourage the reconstruction of the original image $x_0$, it has the possibility to memorize images from the training set. Although the generated diverse samples can qualitatively prove this is not the case, to further quantify the distribution coverage and show the diversity of our approach, we use the improved precision and recall proposed in~\cite{kynkaanniemi2019improved}, and experiment with the unconditional generation task with CelebA-64 and FFHQ-256 dataset with increasing number of training epochs. The results are summarized in Table~\ref{tab:6}.

From Table~\ref{tab:6}, on CelebA-64, it follows that the initial precision is 0.79, and the recall is 0.71. When increasing the number of iterations, the FID remains stable, but the recall starts decreasing from 0.71 to 0.44 at 20 iterations, and it remains stable when further increasing the number of iterations. Similarly, on FFHQ-256 dataset, the initial precision is 0.76, and recall is 0.65. When increasing the iterations, the recall decreases from 0.65 to 0.45 at 20 iterations, and stays more or less the same for increasing numbers of iterations.

Based on the analysis we can see that the diversity is connected to the training epochs of compensation term since it will learn the data distribution to some extent, so we can choose a lower number of epochs during the unconditional generation to increase the diversity of results, and choose a higher number of epochs to generate pixels similar to human faces for face inpainting tasks.





\section{Model architecture} \label{sec:app_architecture}
\begin{figure}[htb]
\centering
\includegraphics[width=\columnwidth]{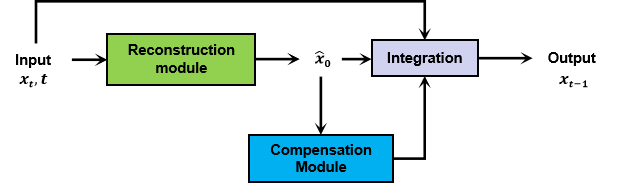}
\caption{\textbf{Schematic overview} of our diffusion model during the denoising process.}
\label{fig:9}
\end{figure}

Our diffusion model is based on the DDIM model, the Ablated Diffusion Model (ADM) (publicly available at \url{https://github.com/openai/guided-diffusion}). We summarize the architecture in Table~\ref{tab:3}, and show the pipeline of our diffusion model in Figure~\ref{fig:9}. During the denoising process, the input noise data $x_t$ first passes through the reconstruction module (with ADM backbone in Table~\ref{tab:3}) to produce the initial clean data reconstruction $\hat{x_0}$. $\hat{x_0}$ is further processed by the compensation module. The output of compensation module and $\hat{x_0}$ is integrated following our compensation sampling algorithm, and outputs $x_{t-1}$. The pipeline iteratively works until the diffusion model outputs the final reconstruction $x_0$.

\begin{table} [tb]
\centering
\resizebox{1.0\linewidth}{!}{
\begin{tabular}{|c|l|c|c|c|}
\hline
&\multirow{2}{*}{Parameter} & CelebA- & CelebA-HQ- & FFHQ-\\
& & 64 & 256 & 256\\
\hline
\multirow{4}{*}{Reconstruction module}& Batch size & 128 & 64 & 64 \\
&Base channels & 64 & 128 & 128 \\
&Channel multipliers  & [1,2,4,8] & [1,1,2,2,4,4] & [1,1,2,2,4,4] \\
&Attention resolution & [16] & [16] & [16] \\
\hline
\multirow{4}{*}{Compensation module}& Batch size & 128 & 64 & 64 \\
&Base channels & 64 & 128 & 128 \\
&Channel multipliers  & [1,2,4,8] & [1,1,2,2,2,2] & [1,1,2,2,2,2] \\

\hline
\end{tabular}
}
\caption{\textbf{Network architecture} of our diffusion model.}
\label{tab:3}
\end{table}

%% file: main.bbl
\begin{thebibliography}{63}
\providecommand{\natexlab}[1]{#1}
\providecommand{\url}[1]{\texttt{#1}}
\expandafter\ifx\csname urlstyle\endcsname\relax
  \providecommand{\doi}[1]{doi: #1}\else
  \providecommand{\doi}{doi: \begingroup \urlstyle{rm}\Url}\fi

\bibitem[Arad~Hudson and Zitnick(2021)]{arad2021compositional}
Dor Arad~Hudson and Larry Zitnick.
\newblock Compositional transformers for scene generation.
\newblock \emph{Advances in Neural Information Processing Systems}, 34:\penalty0 9506--9520, 2021.

\bibitem[Bansal et~al.(2022)Bansal, Borgnia, Chu, Li, Kazemi, Huang, Goldblum, Geiping, and Goldstein]{bansal2022cold}
Arpit Bansal, Eitan Borgnia, Hong-Min Chu, Jie~S Li, Hamid Kazemi, Furong Huang, Micah Goldblum, Jonas Geiping, and Tom Goldstein.
\newblock Cold {D}iffusion: Inverting arbitrary image transforms without noise.
\newblock \emph{arXiv preprint arXiv:2208.09392}, 2022.

\bibitem[Bao et~al.(2022{\natexlab{a}})Bao, Li, Sun, Zhu, and Zhang]{bao2022estimating}
Fan Bao, Chongxuan Li, Jiacheng Sun, Jun Zhu, and Bo Zhang.
\newblock Estimating the optimal covariance with imperfect mean in diffusion probabilistic models.
\newblock \emph{arXiv preprint arXiv:2206.07309}, 2022{\natexlab{a}}.

\bibitem[Bao et~al.(2022{\natexlab{b}})Bao, Li, Zhu, and Zhang]{bao2022analyticdpm}
Fan Bao, Chongxuan Li, Jun Zhu, and Bo Zhang.
\newblock Analytic-{DPM}: An analytic estimate of the optimal reverse variance in diffusion probabilistic models.
\newblock In \emph{International Conference on Learning Representations}, 2022{\natexlab{b}}.

\bibitem[Batzolis et~al.(2021)Batzolis, Stanczuk, Sch{\"o}nlieb, and Etmann]{batzolis2021conditional}
Georgios Batzolis, Jan Stanczuk, Carola-Bibiane Sch{\"o}nlieb, and Christian Etmann.
\newblock Conditional image generation with score-based diffusion models.
\newblock \emph{arXiv preprint arXiv:2111.13606}, 2021.

\bibitem[Cheung et~al.(2021)Cheung, Li, and Zou]{cheung2021facial}
Yiu-Ming Cheung, Mengke Li, and Rong Zou.
\newblock Facial structure guided {GAN} for identity-preserved face image de-occlusion.
\newblock In \emph{International Conference on Multimedia Retrieval}, pages 46--54, 2021.

\bibitem[Choi et~al.(2022)Choi, Lee, Shin, Kim, Kim, and Yoon]{choi2022perception}
Jooyoung Choi, Jungbeom Lee, Chaehun Shin, Sungwon Kim, Hyunwoo Kim, and Sungroh Yoon.
\newblock Perception prioritized training of diffusion models.
\newblock In \emph{IEEE/CVF Conference on Computer Vision and Pattern Recognition}, pages 11472--11481, 2022.

\bibitem[Daras et~al.(2023)Daras, Delbracio, Talebi, Dimakis, and Milanfar]{daras2023soft}
Giannis Daras, Mauricio Delbracio, Hossein Talebi, Alexandros Dimakis, and Peyman Milanfar.
\newblock Soft {D}iffusion: Score matching with general corruptions.
\newblock \emph{Transactions on Machine Learning Research (TMLR)}, 2023.

\bibitem[Dhariwal and Nichol(2021)]{dhariwal2021diffusion}
Prafulla Dhariwal and Alexander Nichol.
\newblock Diffusion models beat {GAN}s on image synthesis.
\newblock \emph{Advances in Neural Information Processing Systems}, 34:\penalty0 8780--8794, 2021.

\bibitem[Dong et~al.(2020)Dong, Zhang, Zhang, and Liu]{dong2020occlusion}
Jiayuan Dong, Liyan Zhang, Hanwang Zhang, and Weichen Liu.
\newblock Occlusion-aware {GAN} for face de-occlusion in the wild.
\newblock In \emph{IEEE International Conference on Multimedia and Expo (ICME)}, pages 1--6, 2020.

\bibitem[Esser et~al.(2021)Esser, Rombach, and Ommer]{esser2021taming}
Patrick Esser, Robin Rombach, and Bjorn Ommer.
\newblock Taming transformers for high-resolution image synthesis.
\newblock In \emph{IEEE/CVF Conference on Computer Vision and Pattern Recognition}, pages 12873--12883, 2021.

\bibitem[Fei et~al.(2023)Fei, Lyu, Pan, Zhang, Yang, Luo, Zhang, and Dai]{fei2023generative}
Ben Fei, Zhaoyang Lyu, Liang Pan, Junzhe Zhang, Weidong Yang, Tianyue Luo, Bo Zhang, and Bo Dai.
\newblock Generative diffusion prior for unified image restoration and enhancement.
\newblock In \emph{IEEE/CVF Conference on Computer Vision and Pattern Recognition}, pages 9935--9946, 2023.

\bibitem[Goodfellow et~al.(2014)Goodfellow, Pouget-Abadie, Mirza, Xu, Warde-Farley, Ozair, Courville, and Bengio]{NIPS2014_5ca3e9b1}
Ian Goodfellow, Jean Pouget-Abadie, Mehdi Mirza, Bing Xu, David Warde-Farley, Sherjil Ozair, Aaron Courville, and Yoshua Bengio.
\newblock Generative adversarial nets.
\newblock In \emph{Advances in Neural Information Processing Systems}, 2014.

\bibitem[Heusel et~al.(2017)Heusel, Ramsauer, Unterthiner, Nessler, and Hochreiter]{heusel2017gans}
Martin Heusel, Hubert Ramsauer, Thomas Unterthiner, Bernhard Nessler, and Sepp Hochreiter.
\newblock {GAN}s trained by a two time-scale update rule converge to a local {N}ash equilibrium.
\newblock \emph{Advances in neural information processing systems}, 30, 2017.

\bibitem[Ho et~al.(2020)Ho, Jain, and Abbeel]{ho2020denoising}
Jonathan Ho, Ajay Jain, and Pieter Abbeel.
\newblock Denoising diffusion probabilistic models.
\newblock \emph{Advances in Neural Information Processing Systems}, 33:\penalty0 6840--6851, 2020.

\bibitem[Jain et~al.(2023)Jain, Zhou, Yu, and Shi]{jain2023keys}
Jitesh Jain, Yuqian Zhou, Ning Yu, and Humphrey Shi.
\newblock Keys to better image inpainting: Structure and texture go hand in hand.
\newblock In \emph{IEEE/CVF Winter Conference on Applications of Computer Vision}, pages 208--217, 2023.

\bibitem[Karras et~al.(2018)Karras, Aila, Laine, and Lehtinen]{karras2018progressive}
Tero Karras, Timo Aila, Samuli Laine, and Jaakko Lehtinen.
\newblock Progressive growing of {GAN}s for improved quality, stability, and variation.
\newblock In \emph{International Conference on Learning Representations}, 2018.

\bibitem[Karras et~al.(2019)Karras, Laine, and Aila]{karras2019style}
Tero Karras, Samuli Laine, and Timo Aila.
\newblock A style-based generator architecture for generative adversarial networks.
\newblock In \emph{IEEE/CVF Conference on Computer Vision and Pattern Recognition}, pages 4401--4410, 2019.

\bibitem[Karras et~al.(2020)Karras, Laine, Aittala, Hellsten, Lehtinen, and Aila]{karras2020analyzing}
Tero Karras, Samuli Laine, Miika Aittala, Janne Hellsten, Jaakko Lehtinen, and Timo Aila.
\newblock Analyzing and improving the image quality of {StyleGAN}.
\newblock In \emph{IEEE/CVF Conference on Computer Vision and Pattern Recognition}, pages 8110--8119, 2020.

\bibitem[Karras et~al.(2022)Karras, Aittala, Aila, and Laine]{karras2022elucidating}
Tero Karras, Miika Aittala, Timo Aila, and Samuli Laine.
\newblock Elucidating the design space of diffusion-based generative models.
\newblock \emph{arXiv preprint arXiv:2206.00364}, 2022.

\bibitem[Kim et~al.(2021)Kim, Shin, Song, Kang, and Moon]{kim2021soft}
Dongjun Kim, Seungjae Shin, Kyungwoo Song, Wanmo Kang, and Il-Chul Moon.
\newblock Soft {T}runcation: A universal training technique of score-based diffusion model for high precision score estimation.
\newblock \emph{arXiv preprint arXiv:2106.05527}, 2021.

\bibitem[Kingma and Welling(2013)]{kingma2013auto}
Diederik~P Kingma and Max Welling.
\newblock Auto-encoding variational bayes.
\newblock \emph{arXiv preprint arXiv:1312.6114}, 2013.

\bibitem[Krizhevsky(2009)]{krizhevsky2009cifar10}
Alex Krizhevsky.
\newblock Learning multiple layers of features from tiny images.
\newblock Technical report, University of Toronto, 2009.

\bibitem[Kynk{\"a}{\"a}nniemi et~al.(2019)Kynk{\"a}{\"a}nniemi, Karras, Laine, Lehtinen, and Aila]{kynkaanniemi2019improved}
Tuomas Kynk{\"a}{\"a}nniemi, Tero Karras, Samuli Laine, Jaakko Lehtinen, and Timo Aila.
\newblock Improved precision and recall metric for assessing generative models.
\newblock \emph{Advances in Neural Information Processing Systems}, 32, 2019.

\bibitem[Li et~al.(2022{\natexlab{a}})Li, Zheng, Wang, Yao, Chen, Ding, and Li]{li2022entropy}
Shengming Li, Guangcong Zheng, Hui Wang, Taiping Yao, Yang Chen, Shoudong Ding, and Xi Li.
\newblock Entropy-driven sampling and training scheme for conditional diffusion generation.
\newblock \emph{arXiv preprint arXiv:2206.11474}, 2022{\natexlab{a}}.

\bibitem[Li et~al.(2022{\natexlab{b}})Li, Lin, Zhou, Qi, Wang, and Jia]{li2022mat}
Wenbo Li, Zhe Lin, Kun Zhou, Lu Qi, Yi Wang, and Jiaya Jia.
\newblock {MAT}: Mask-aware transformer for large hole image inpainting.
\newblock In \emph{IEEE/CVF Conference on Computer Vision and Pattern Recognition}, pages 10758--10768, 2022{\natexlab{b}}.

\bibitem[Li et~al.(2023)Li, Liu, Lian, Yang, Dong, Kang, Zhang, and Keutzer]{Li_2023_ICCV}
Xiuyu Li, Yijiang Liu, Long Lian, Huanrui Yang, Zhen Dong, Daniel Kang, Shanghang Zhang, and Kurt Keutzer.
\newblock Q-diffusion: Quantizing diffusion models.
\newblock In \emph{IEEE/CVF International Conference on Computer Vision}, pages 17535--17545, 2023.

\bibitem[Liu et~al.(2022{\natexlab{a}})Liu, Ren, Lin, and Zhao]{liu2022pseudo}
Luping Liu, Yi Ren, Zhijie Lin, and Zhou Zhao.
\newblock Pseudo numerical methods for diffusion models on manifolds.
\newblock \emph{arXiv preprint arXiv:2202.09778}, 2022{\natexlab{a}}.

\bibitem[Liu et~al.(2022{\natexlab{b}})Liu, Gong, and Liu]{liu2022flow}
Xingchao Liu, Chengyue Gong, and Qiang Liu.
\newblock Flow straight and fast: Learning to generate and transfer data with rectified flow.
\newblock \emph{arXiv preprint arXiv:2209.03003}, 2022{\natexlab{b}}.

\bibitem[Liu et~al.(2015)Liu, Luo, Wang, and Tang]{liu2015deep}
Ziwei Liu, Ping Luo, Xiaogang Wang, and Xiaoou Tang.
\newblock Deep learning face attributes in the wild.
\newblock In \emph{IEEE International Conference on Computer Vision}, pages 3730--3738, 2015.

\bibitem[Lu et~al.(2022{\natexlab{a}})Lu, Zhou, Bao, Chen, Li, and Zhu]{lu2022dpm}
Cheng Lu, Yuhao Zhou, Fan Bao, Jianfei Chen, Chongxuan Li, and Jun Zhu.
\newblock {DPM-S}olver: A fast ode solver for diffusion probabilistic model sampling in around 10 steps.
\newblock \emph{Advances in Neural Information Processing Systems}, 35:\penalty0 5775--5787, 2022{\natexlab{a}}.

\bibitem[Lu et~al.(2022{\natexlab{b}})Lu, Jiang, Huang, Wu, and Liu]{lu2022glama}
Zeyu Lu, Junjun Jiang, Junqin Huang, Gang Wu, and Xianming Liu.
\newblock {GLaMa}: Joint spatial and frequency loss for general image inpainting.
\newblock In \emph{IEEE/CVF Conference on Computer Vision and Pattern Recognition}, pages 1301--1310, 2022{\natexlab{b}}.

\bibitem[Lugmayr et~al.(2022)Lugmayr, Danelljan, Romero, Yu, Timofte, and Van~Gool]{lugmayr2022repaint}
Andreas Lugmayr, Martin Danelljan, Andres Romero, Fisher Yu, Radu Timofte, and Luc Van~Gool.
\newblock Repaint: Inpainting using denoising diffusion probabilistic models.
\newblock In \emph{IEEE/CVF Conference on Computer Vision and Pattern Recognition}, pages 11461--11471, 2022.

\bibitem[Mahendran and Vedaldi(2015)]{mahendran2015understanding}
Aravindh Mahendran and Andrea Vedaldi.
\newblock Understanding deep image representations by inverting them.
\newblock In \emph{IEEE Conference on Computer Vision and Pattern Recognition}, pages 5188--5196, 2015.

\bibitem[Meng et~al.(2023)Meng, Rombach, Gao, Kingma, Ermon, Ho, and Salimans]{Meng_2023_CVPR}
Chenlin Meng, Robin Rombach, Ruiqi Gao, Diederik Kingma, Stefano Ermon, Jonathan Ho, and Tim Salimans.
\newblock On distillation of guided diffusion models.
\newblock In \emph{IEEE/CVF Conference on Computer Vision and Pattern Recognition}, pages 14297--14306, 2023.

\bibitem[Moghadam et~al.(2023)Moghadam, Van~Dalen, Martin, Lennerz, Yip, Farahani, and Bashashati]{moghadam2023morphology}
Puria~Azadi Moghadam, Sanne Van~Dalen, Karina~C Martin, Jochen Lennerz, Stephen Yip, Hossein Farahani, and Ali Bashashati.
\newblock A morphology focused diffusion probabilistic model for synthesis of histopathology images.
\newblock In \emph{IEEE/CVF Winter Conference on Applications of Computer Vision}, pages 2000--2009, 2023.

\bibitem[Nash et~al.(2021)Nash, Menick, Dieleman, and Battaglia]{nash2021generating}
Charlie Nash, Jacob Menick, Sander Dieleman, and Peter~W Battaglia.
\newblock Generating images with sparse representations.
\newblock \emph{arXiv preprint arXiv:2103.03841}, 2021.

\bibitem[Pandey et~al.(2022)Pandey, Mukherjee, Rai, and Kumar]{pandey2022diffusevae}
Kushagra Pandey, Avideep Mukherjee, Piyush Rai, and Abhishek Kumar.
\newblock {DiffuseVAE}: Efficient, controllable and high-fidelity generation from low-dimensional latents.
\newblock \emph{arXiv preprint arXiv:2201.00308}, 2022.

\bibitem[Preechakul et~al.(2022)Preechakul, Chatthee, Wizadwongsa, and Suwajanakorn]{preechakul2022diffusion}
Konpat Preechakul, Nattanat Chatthee, Suttisak Wizadwongsa, and Supasorn Suwajanakorn.
\newblock Diffusion autoencoders: Toward a meaningful and decodable representation.
\newblock In \emph{IEEE/CVF Conference on Computer Vision and Pattern Recognition}, pages 10619--10629, 2022.

\bibitem[Ramesh et~al.(2022)Ramesh, Dhariwal, Nichol, Chu, and Chen]{ramesh2022hierarchical}
Aditya Ramesh, Prafulla Dhariwal, Alex Nichol, Casey Chu, and Mark Chen.
\newblock Hierarchical text-conditional image generation with clip latents.
\newblock \emph{arXiv preprint arXiv:2204.06125}, 2022.

\bibitem[Ronneberger et~al.(2015)Ronneberger, Fischer, and Brox]{ronneberger2015u}
Olaf Ronneberger, Philipp Fischer, and Thomas Brox.
\newblock U-net: Convolutional networks for biomedical image segmentation.
\newblock In \emph{Medical Image Computing and Computer-Assisted Intervention--MICCAI 2015: 18th International Conference, Munich, Germany, October 5-9, 2015, Proceedings, Part III 18}, pages 234--241, 2015.

\bibitem[Schonfeld et~al.(2020)Schonfeld, Schiele, and Khoreva]{schonfeld2020u}
Edgar Schonfeld, Bernt Schiele, and Anna Khoreva.
\newblock A {U}-net based discriminator for generative adversarial networks.
\newblock In \emph{IEEE/CVF Conference on Computer Vision and Pattern Recognition}, pages 8207--8216, 2020.

\bibitem[Sinha et~al.(2021)Sinha, Song, Meng, and Ermon]{sinha2021d2c}
Abhishek Sinha, Jiaming Song, Chenlin Meng, and Stefano Ermon.
\newblock {D2C}: Diffusion-decoding models for few-shot conditional generation.
\newblock \emph{Advances in Neural Information Processing Systems}, 34:\penalty0 12533--12548, 2021.

\bibitem[Sohl-Dickstein et~al.(2015)Sohl-Dickstein, Weiss, Maheswaranathan, and Ganguli]{sohl2015deep}
Jascha Sohl-Dickstein, Eric Weiss, Niru Maheswaranathan, and Surya Ganguli.
\newblock Deep unsupervised learning using nonequilibrium thermodynamics.
\newblock In \emph{International Conference on Machine Learning}, pages 2256--2265, 2015.

\bibitem[Song et~al.(2021)Song, Meng, and Ermon]{song2021denoising}
Jiaming Song, Chenlin Meng, and Stefano Ermon.
\newblock Denoising diffusion implicit models.
\newblock In \emph{International Conference on Learning Representations}, 2021.

\bibitem[Song and Ermon(2019)]{song2019generative}
Yang Song and Stefano Ermon.
\newblock Generative modeling by estimating gradients of the data distribution.
\newblock \emph{Advances in Neural Information Processing Systems}, 32, 2019.

\bibitem[Song et~al.(2020)Song, Sohl-Dickstein, Kingma, Kumar, Ermon, and Poole]{song2020score}
Yang Song, Jascha Sohl-Dickstein, Diederik~P Kingma, Abhishek Kumar, Stefano Ermon, and Ben Poole.
\newblock Score-based generative modeling through stochastic differential equations.
\newblock \emph{arXiv preprint arXiv:2011.13456}, 2020.

\bibitem[Such et~al.(2020)Such, Rawal, Lehman, Stanley, and Clune]{such2020generative}
Felipe~Petroski Such, Aditya Rawal, Joel Lehman, Kenneth Stanley, and Jeffrey Clune.
\newblock Generative teaching networks: Accelerating neural architecture search by learning to generate synthetic training data.
\newblock In \emph{International Conference on Machine Learning}, pages 9206--9216, 2020.

\bibitem[Suvorov et~al.(2022)Suvorov, Logacheva, Mashikhin, Remizova, Ashukha, Silvestrov, Kong, Goka, Park, and Lempitsky]{suvorov2022resolution}
Roman Suvorov, Elizaveta Logacheva, Anton Mashikhin, Anastasia Remizova, Arsenii Ashukha, Aleksei Silvestrov, Naejin Kong, Harshith Goka, Kiwoong Park, and Victor Lempitsky.
\newblock Resolution-robust large mask inpainting with {F}ourier convolutions.
\newblock In \emph{IEEE/CVF Winter Conference on Applications of Computer Vision}, pages 2149--2159, 2022.

\bibitem[Vahdat and Kautz(2020)]{vahdat2020nvae}
Arash Vahdat and Jan Kautz.
\newblock {NVAE}: A deep hierarchical variational autoencoder.
\newblock \emph{Advances in Neural Information Processing Systems}, 33:\penalty0 19667--19679, 2020.

\bibitem[Vahdat et~al.(2021)Vahdat, Kreis, and Kautz]{vahdat2021score}
Arash Vahdat, Karsten Kreis, and Jan Kautz.
\newblock Score-based generative modeling in latent space.
\newblock \emph{Advances in Neural Information Processing Systems}, 34:\penalty0 11287--11302, 2021.

\bibitem[Wang et~al.(2004)Wang, Bovik, Sheikh, and Simoncelli]{wang2004image}
Zhou Wang, Alan~C Bovik, Hamid~R Sheikh, and Eero~P Simoncelli.
\newblock Image quality assessment: From error visibility to structural similarity.
\newblock \emph{IEEE Transactions on Image Processing}, 13\penalty0 (4):\penalty0 600--612, 2004.

\bibitem[Wang et~al.(2023{\natexlab{a}})Wang, Jiang, Zheng, Wang, He, Wang, Chen, and Zhou]{wang2023patch}
Zhendong Wang, Yifan Jiang, Huangjie Zheng, Peihao Wang, Pengcheng He, Zhangyang Wang, Weizhu Chen, and Mingyuan Zhou.
\newblock Patch diffusion: Faster and more data-efficient training of diffusion models.
\newblock \emph{arXiv preprint arXiv:2304.12526}, 2023{\natexlab{a}}.

\bibitem[Wang et~al.(2023{\natexlab{b}})Wang, Zheng, He, Chen, and Zhou]{wang2023diffusiongan}
Zhendong Wang, Huangjie Zheng, Pengcheng He, Weizhu Chen, and Mingyuan Zhou.
\newblock Diffusion-{GAN}: Training {GAN}s with diffusion.
\newblock In \emph{International Conference on Learning Representations}, 2023{\natexlab{b}}.

\bibitem[Xiao et~al.(2021)Xiao, Kreis, and Vahdat]{xiao2021tackling}
Zhisheng Xiao, Karsten Kreis, and Arash Vahdat.
\newblock Tackling the generative learning trilemma with denoising diffusion {GAN}s.
\newblock \emph{arXiv preprint arXiv:2112.07804}, 2021.

\bibitem[Xu et~al.(2023{\natexlab{a}})Xu, Liu, Tian, Tong, Tegmark, and Jaakkola]{xu2023pfgm++}
Yilun Xu, Ziming Liu, Yonglong Tian, Shangyuan Tong, Max Tegmark, and Tommi Jaakkola.
\newblock {PFGM++}: Unlocking the potential of physics-inspired generative models.
\newblock \emph{arXiv preprint arXiv:2302.04265}, 2023{\natexlab{a}}.

\bibitem[Xu et~al.(2023{\natexlab{b}})Xu, Tong, and Jaakkola]{xu2023stable}
Yilun Xu, Shangyuan Tong, and Tommi~S. Jaakkola.
\newblock Stable target field for reduced variance score estimation in diffusion models.
\newblock In \emph{International Conference on Learning Representations}, 2023{\natexlab{b}}.

\bibitem[Yin et~al.(2023)Yin, Huang, Fu, Wang, and Chen]{yin2023segmentation}
Xiangnan Yin, Di Huang, Zehua Fu, Yunhong Wang, and Liming Chen.
\newblock Segmentation-reconstruction-guided facial image de-occlusion.
\newblock In \emph{IEEE International Conference on Automatic Face and Gesture Recognition (FG)}, pages 1--8, 2023.

\bibitem[Yu et~al.(2018)Yu, Lin, Yang, Shen, Lu, and Huang]{yu2018generative}
Jiahui Yu, Zhe Lin, Jimei Yang, Xiaohui Shen, Xin Lu, and Thomas~S Huang.
\newblock Generative image inpainting with contextual attention.
\newblock In \emph{IEEE Conference on Computer Vision and Pattern Recognition}, pages 5505--5514, 2018.

\bibitem[Zhang et~al.(2022)Zhang, Liu, Han, Wan, and Shao]{zhang2022face}
Ni Zhang, Nian Liu, Junwei Han, Kaiyuan Wan, and Ling Shao.
\newblock Face de-occlusion with deep cascade guidance learning.
\newblock \emph{IEEE Transactions on Multimedia}, 2022.

\bibitem[Zhang et~al.(2018)Zhang, Isola, Efros, Shechtman, and Wang]{zhang2018unreasonable}
Richard Zhang, Phillip Isola, Alexei~A Efros, Eli Shechtman, and Oliver Wang.
\newblock The unreasonable effectiveness of deep features as a perceptual metric.
\newblock In \emph{IEEE Conference on Computer Vision and Pattern Recognition}, pages 586--595, 2018.

\bibitem[Zhao et~al.(2021)Zhao, Cui, Sheng, Dong, Liang, Chang, and Xu]{zhao2021large}
Shengyu Zhao, Jonathan Cui, Yilun Sheng, Yue Dong, Xiao Liang, Eric~I Chang, and Yan Xu.
\newblock Large scale image completion via co-modulated generative adversarial networks.
\newblock \emph{arXiv preprint arXiv:2103.10428}, 2021.

\bibitem[Zhu et~al.(2017)Zhu, Park, Isola, and Efros]{zhu2017unpaired}
Jun-Yan Zhu, Taesung Park, Phillip Isola, and Alexei~A Efros.
\newblock Unpaired image-to-image translation using cycle-consistent adversarial networks.
\newblock In \emph{IEEE International Conference on Computer Vision}, pages 2223--2232, 2017.

\end{thebibliography}
